\documentclass[10pt,journal,compsoc]{IEEEtran}
\usepackage{enumitem}
\usepackage{multirow}
\usepackage{booktabs, tabularx}
\ifCLASSOPTIONcompsoc
  \usepackage[nocompress]{cite}
\else
  \usepackage{cite}
\fi
\usepackage{amsmath}
\usepackage{tikz}
\usetikzlibrary{arrows,shapes,positioning,shadows,trees}
\tikzset{
  basic/.style  = {draw, rounded corners, text width=17mm, align=center,
                   anchor=north, drop shadow, font=\sffamily, rectangle},
  root/.style   = {basic, rounded corners=2pt, thin, align=center,
                   fill=green!55},
  level 2/.style = {basic, rounded corners=7pt, thin,align=center, fill=yellow!70,
                   },
  level 3/.style = {basic,  thin, rounded corners=9pt, align=center, fill=blue!35, yshift=-50pt}
}
\usepackage[edges]{forest}
\usepackage{xcolor}

\newlength{\Lnote}

  \usepackage[]{graphicx}
\usepackage{amsmath}
\usepackage{amsfonts}
\usepackage{array}

\ifCLASSOPTIONcompsoc
 \usepackage[caption=false,font=footnotesize,labelfont=sf,textfont=sf]{subfig}
\else
 \usepackage[caption=false,font=footnotesize]{subfig}
\fi
\usepackage{url}

\usepackage{xr}
\makeatletter

\newcommand*{\addFileDependency}[1]{
\typeout{(#1)}
\@addtofilelist{#1}
\IfFileExists{#1}{}{\typeout{No file #1.}}
}\makeatother

\newcommand*{\myexternaldocument}[1]{%
\externaldocument{#1}%
\addFileDependency{#1.tex}%
\addFileDependency{#1.aux}%
}

\myexternaldocument{supplemental}

\begin{document}
%
\title{The Neural Process Family: Survey, Applications and Perspectives}
%
%
%
%

\author{Saurav~Jha,~
        Dong~Gong, 
        ~Xuesong~Wang,
        ~Richard~Turner,
        ~Lina~Yao
\IEEEcompsocitemizethanks{\IEEEcompsocthanksitem Saurav Jha and Dong Gong are with the School of Computer Science and Engineering, UNSW Sydney, Australia.\protect\\
E-mail: saurav.jha@unsw.edu.au
\IEEEcompsocthanksitem Richard Turner is with the Department of Engineering, University of Cambridge, UK.
\IEEEcompsocthanksitem Lina Yao and Xuesong Wang are with CSIRO's Data61 and the School of Computer Science and Engineering, UNSW Sydney, Australia.}
}

%
%

\markboth{The Neural Process Family: Survey, Applications and Perspectives}%
{Jha \MakeLowercase{\textit{et al.}}: Bare Demo of IEEEtran.cls for Computer Society Journals}
%



\IEEEtitleabstractindextext{%
\begin{abstract}
The standard approaches to neural network implementation yield powerful function approximation capabilities but are limited in their abilities to learn meta-representations and to reason  uncertainties in their predictions. Gaussian processes, on the other hand, adopt the Bayesian learning scheme to estimate such uncertainties but are constrained by their efficiency and approximation capacity. The Neural Processes Family (NPF) intends to offer the best of both worlds by leveraging neural networks for meta-learning predictive uncertainties. Such potential has brought substantial research activity to the family in recent years. A comprehensive survey of NPF models is thus needed to organize and relate their motivation, methodology, and experiments. This paper intends to address this gap while digging  into the formulation, research themes, and applications concerning the NPF members. We shed light on their potential to bring several recent advances in  deep learning domains under one umbrella. We then provide a rigorous taxonomy of the family and empirically demonstrate their capabilities for modeling data generating functions operating on 1-d, 2-d, and 3-d input domains.  We conclude by discussing our perspectives on the promising directions that can fuel the research advances in the field.\end{abstract}

}

\maketitle

\IEEEdisplaynontitleabstractindextext

%
\IEEEpeerreviewmaketitle

\IEEEraisesectionheading{\section{Introduction}\label{sec:introduction}}

%
%
%
%
Real-world applications of deep learning models often involve industrial and safety critical decision-making areas including but not limited to autonomous driving \cite{casas2020implicit} and navigation \cite{abcouwer2021machine}, weather prediction and climate modeling \cite{quinting2022eulerian}, and disease \cite{mcbee2018deep} and workplace hazard \cite{wang2019predicting} detection.    A necessary trait for models in these environments is the ability to capture the  epistemic uncertainty in their predictions arising from the lack of knowledge of the underlying data generating processes. Uncertainty awareness thus counts among the major prerequisites for practical machine learning models today \cite{yanlecun2022}. Consequently,  recent advances in deep learning have seen  designs that can better  measure uncertainties in their predictions \cite{louart2018random}, mitigate the risks attached with their tendency to be overconfident about   predictions \cite{pereira2020challenges}, all while being able to meta learn the uncertainties \cite{nguyen2020uncertainty}. 

For the aforesaid objectives, the Bayesian paradigm  \cite{mackay1995probable}, with its ability to encode a prior distribution on the parameters $\theta$ of the neural network, stands among the earliest promising frameworks. Upon the availability of more data, such priors can in turn be employed to periodically influence the posterior distribution. 
Nevertheless, Bayesian Neural Networks (BNNs) have three fundamental issues inherent to their nature. First, it is hard to specify generalization-sensitive weight priors in BNNs, \textit{i.e.,} priors that assign higher likelihoods to better posteriors. The absence of such priors takes away the guarantee that the output distributions of BNNs reflect the true posteriors and hence,  the true uncertainties in predictions \cite{blogpost}. Second, the inference techniques  approximating the true posterior in BNNs are in general, not scalable to high dimensional probability models and large datasets. Third and most fundamental, the motivation behind going Bayesian at the first place may not always be clear for a range of real-world problems.\footnote{https://mlg-blog.com/2021/03/31/what-keeps-a-bayesian-awake-at-night-part-2.html}
\par
\begin{figure}[t!]
    \centering
      
        \includegraphics[width=0.5\textwidth]{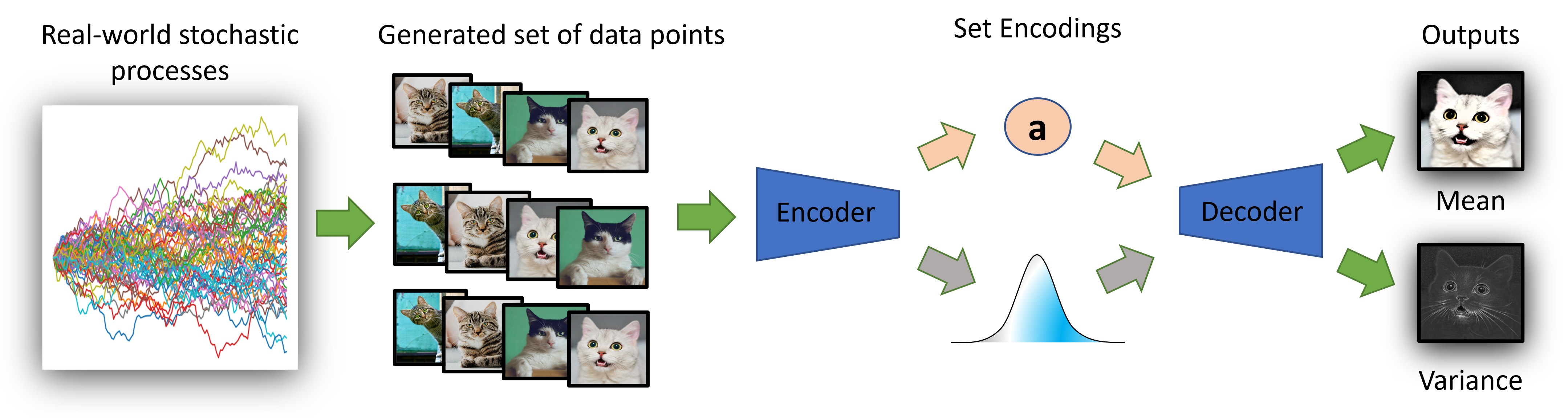}

\caption{An illustration of a neural process modeling the stochasticities underlying data to generate predictive mean and uncertainties. \textcolor{orange}{Orange} denotes the deterministic path while \textcolor{gray}{Gray} denotes the latent path.}
\label{fig:intro}%
\end{figure}
 \textbf{The Neural Process Family.} The uncertainty-aware Neural Process Family (NPF) \cite{garnelo2018conditional} aims to address the aforementioned limitations of the Bayesian paradigm by exploiting the function approximation capabilities of deep neural networks to learn a family of real-world data-generating processes, a.k.a., stochastic Gaussian  Processes (GPs) \cite{williams2006gaussian}.\footnote{To keep our  discussion centered around deep learners, we brief GPs and their limitations in Appendix \ref{app:gp}.} NPs define uncertainties in predictions in terms of a conditional distribution over functions  where each function $f$ is parameterized using a neural network.
To model the variability of $f$ based on the variability of the generated data, NPs concurrently train and test their learned parameters on multiple datasets. This endows them with the capability to meta learn their predictive distributions over functions.  NPF members thus combine the best of meta learners, GPs and neural networks. Like GPs, NPs learn a distribution of functions, quickly adapt to new observations, and provide uncertainty measures given test time observations. Like neural networks, NPs learn function approximation from data directly, besides being efficient at inference. 

 NPs incorporate the encoder-decoder architecture that comprises a functional encoding of each observation point followed by the learning of a decoder function whose parameters are capable of unraveling the unobserved function realizations to approximate the outputs of $f$ (see fig. \ref{fig:intro}). Despite their resemblance to NPs, the vanilla encoder-decoder networks traditionally based on CNNs, RNNs, and Transformers \cite{vaswani2017attention} operate merely on pointwise inputs and clearly lack the incentive to meta learn representations  for dynamically changing functions (imagine $f$ changing over a continuum such as time) and their families. The NPF members  leverage these architectures to model functional input spaces and provide uncertainty-aware estimates.

\textbf{The motivation for this survey.} 
We conduct this survey for two main reasons. First and foremost, there exists no work detailing a comprehensive overview of the progress since the dawn of the NP family. However, the field has gathered $50+$ papers over the years while incorporating a range of deep learning  advances. Therefore, a broader survey is needed to track the scope of work in the domain actively. Second, we find that although NPs were originally introduced for regression, they have so far been employed on a number of topics requiring the measure of uncertainty. 

The  diversity of topics  with the absolute absence of a formal literature review presents a number of challenges. How can we effectively cover all existing advances in the Neural Process Family? On what basis do we group the existing works? How do we relate these advances to other deep learning domains and/or the first principles that they are based upon? How can we  give the readers a taste of fundamental implementation  such as the form of the input-output mapping functions represented by the NPF architectures?  Finally, what existing limitations do we analyze that can propel long-term research impacts in the field?

\textbf{Our contributions.} To answer the above questions, we first lay the foundations of the family and relate these from the eye of a number of deep learning domains - including but not limited to deep learning on sets, functions, and graphs, generative modeling, and Bayesian deep learning - whose combined advances are leveraged by the NPF members. We then study the use of inductive biases as a comprehensive classification basis for the existing NPF members. Besides  providing an in-depth classification of these, we also explicitly elaborate the application domains and wherever applicable, the type of tasks within each domain that these can handle. To give a clearer picture of the latter, we consider visualizing the function modeling of some most prominent NPF architectures.  
Our key contributions are:

\begin{enumerate}[leftmargin=0.4cm]
    \item We present a detailed taxonomy of advancements in the Neural Process Family based on their primarily targeted inductive biases. We further discuss the  family from the viewpoints of other deep learning domains. 

    \item 
    There exist a parallel direction of the NPF research focusing on adapting the foundations mentioned in the taxonomy to application-specific purposes. We coin this line of research as application-specific NPF advances  and detail six such major application domains.
    
    \item To sketch how NPs  accomplish different types of tasks, 
    we visualize and discuss the 1-d, 2-d and 3-d function modeling results of common NPF members. 
    
    \item We identify the limitations in the current NPF branches and discuss perspectives on several such directions that are plausible of bringing far-sighted impacts to the field.
\end{enumerate}


The rest of the paper is organized as follows. Section \ref{sec:background} defines the problem setup and the preliminaries on the groundbreaking works. Section \ref{sec:resemblances} draws connections of NPs with a range of deep learning domains. We begin our survey of the  NP family by first presenting a comprehensive taxonomy of these in section  \ref{sec:classification}.  In section \ref{sec:applications}, we outline the application-specific advances in the family. Sections \ref{sec:datasets} and \ref{sec:task_nature} then discuss the domain areas and the  functions that NPs commonly tackle.  Finally, section \ref{sec:future_dir} compiles a list of our perspectives on  the future research directions.


\section{Background}
\label{sec:background}
\textbf{Problem Setup:} We are interested in modeling a real-world stochastic process as a data-generating function $f: \mathcal{X} \rightarrow \mathcal{Y}$ that is continuous, bounded, and random.
 We define a dataset $D \sim f$ to be composed of a labeled context set $C = (X_C, Y_C) = \{(x_i, y_i)\}_{i=1}^{N} $ with inputs $x_i \in \mathcal{X}$ (function locations) and outputs $y_i \in \mathcal{Y}$ (function values), and an unlabeled target set $T = X_T = \{x_i\}_{i=1}^{N+M} \in \mathcal{X}$ ($C$ and $T$ thus follow the same distribution \cite{le2018empirical}).  We assume the location and values to be finite dimensional, \textit{i.e.}, $x_i \in \mathbb{R}^{d_x}, y_i \in \mathbb{R}^{d_y}$. To learn $f$ using $D$, we try to mimic the conditional distribution $p(f(T)| C, T)$ which in turn is a joint distribution over the random variables $\{f(x_i)\}_{i=1}^{N+M}$. Training on a finite set of all observable locations of multiple such datasets thus  helps capture the ground truth stochastic process with the underlying distribution $\mathcal{D}$.%
\par
Given $C$ and $T$, our  learning objective boils down to finding the model with the optimal parameters $\theta^*$ that maximize the likelihood of the predictive distribution $p(f(T)|T, C;,\theta)$. The statistics (mean and variance) of this distribution  provides an estimation of how uncertain the trained model is regarding its predictions mimicking the ground truth values. Accordingly, one can leverage a meta-learning setup where, a neural network with parameters $\theta$ is concurrently trained and tested on multiple datasets $\{(X^i_{C \cup T}, Y^i_C)\}_{i=1}^n \sim D_i$ 
to capture the variability of  $f$
based on the variability of the training datasets. On these premises, Table \ref{tab:TableOfNotationForMyResearch} lists the notations used throughout this paper.

\begin{table}[h!]\caption{An index of notations used throughout the paper}
\small
\centering 
\begin{tabular}{r  p{6.5cm} }
\toprule
 \multicolumn{2}{c}{\underline{\textbf{Function variables}}}\\
\multicolumn{2}{c}{}\\
$f: \mathcal{X} \rightarrow \mathcal{Y}$ &  function modeling a stochastic process. \\
    $\phi$, $\rho$ & the set encoder and decoder, respectively. \\
  $a$ & the permutation-invariant  operation. \\
  $E$ &  an encoding function: $E = a \circ \phi$. \\ 
    $\Phi$,  $\Phi_{conv}$, $\Phi_{eq}$ & the Deep Set \cite{zaheer2017deep}: $\Phi = \rho \circ E $, the ConvDeepSet \cite{Gordon2020Convolutional}, and  the EquivDeepSet \cite{kawano2021group}, respectively.\\
\multicolumn{2}{c}{\underline{\textbf{Data Variables}}}\\
\multicolumn{2}{c}{}\\
 $\mathcal{X}$, $\mathcal{Y}$ & the domain and the range of $f$, respectively. \\ 
    $D$ & a dataset generated by $f$: $D = (\mathcal{X}, \mathcal{Y})$.\\ 
  $x, y$ & a real-valued data and target point: $x \in \mathbb{R}^{d_x}, x \in \mathcal{X}$; $y \in \mathbb{R}^{d_y}$, $y \in \mathcal{Y}$.  \\
  $C$ & the context set: $C = \{(x_i, y_i)\}_{i=1}^N$. \\
  $T$ & the target set: $T = \{x_i\}_{i = 1}^{M+N}$.  \\
      $N, M$ & the cardinalities of $C$ and $T$, respectively. \\ 
  $\theta$ & the parameter space of $\phi \circ \rho$.\\
    $S$ & total time steps of a sequence of tasks. \\
    $t$ & an instant among the total time steps, $t \in S$. \\
    $\mathcal{T}$ & a meta-learning task on samples of $D$. \\
    \multicolumn{2}{c}{\underline{\textbf{Probability distributions}}}\\
\multicolumn{2}{c}{}\\
 $p$ & the true probability distribution. \\
    $q$ & the variational approximation to $p$. \\
    $\mathcal{D}$ & the distribution of $f$ that is to be meta-learned. \\
 \bottomrule
\end{tabular}
\label{tab:TableOfNotationForMyResearch}
\end{table}

\subsection{Preliminaries for the Neural Process Family}
\label{sec:preliminaries}
In this section, we describe the two seminal works laying the foundation for modeling stochastic processes using neural networks. These are preliminaries for the Neural Process Familty (NPF) as all subsequent works build upon these.

\begin{figure*}[t!]
\centering
\subfloat[Conditional Neural Process (CNP)]{\includegraphics[width=2in]{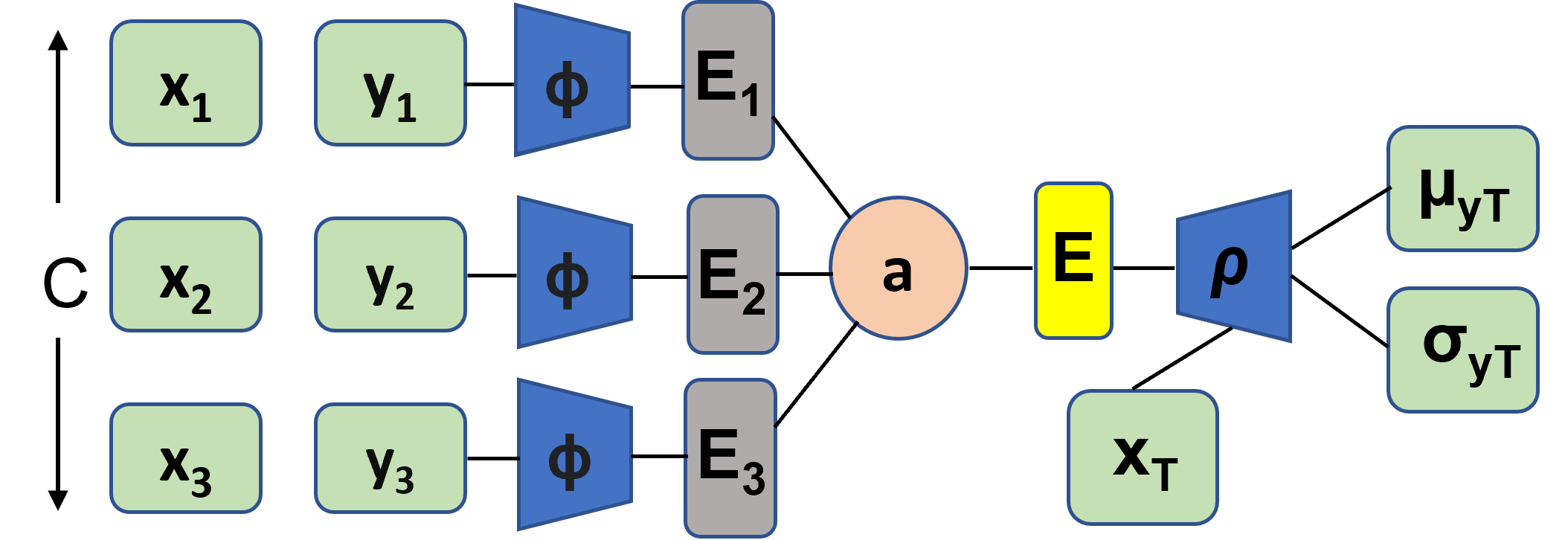}%
 \label{fig:cata}
}
\hspace{0.5in}
\subfloat[(Latent) Neural Process (NP)]{\includegraphics[width=2.2in]{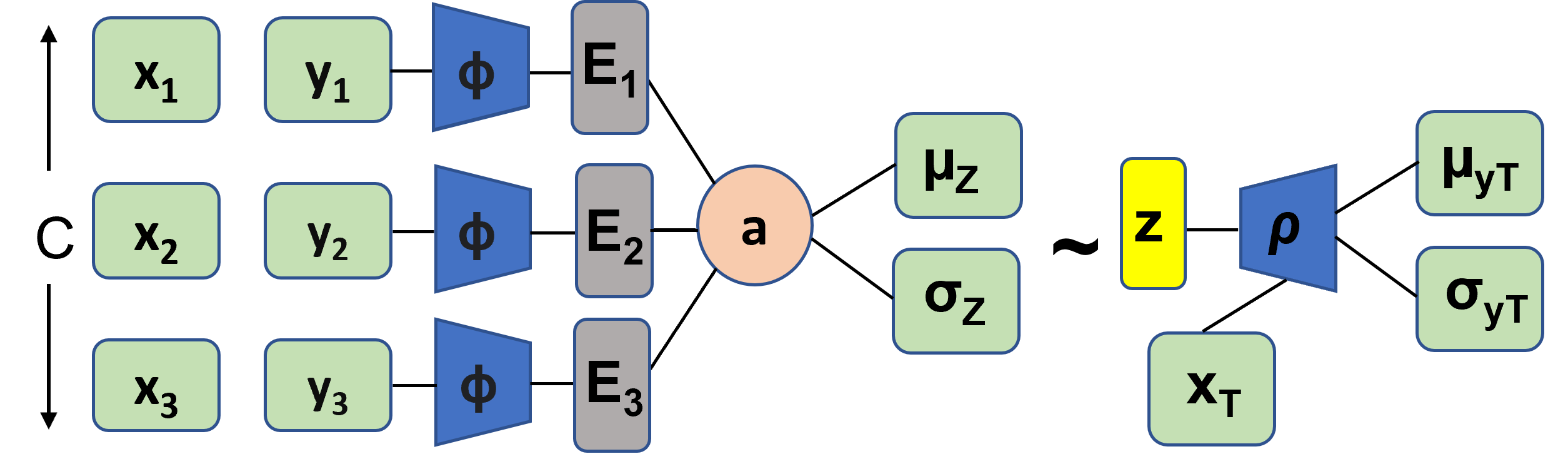}%
 \label{fig:catb}
}

\caption{Illustration of Neural Process architectures adapted from \cite{garnelo2018conditional}: $C$ is the context set composed of three labeled data points $(x_i, y_i)$. $\phi$ is the encoder network acting on individual data points to produce encodings  $E_i$ which are aggregated by the operator $a$. $\rho$ is the decoder network conditioned on the target location $x_T$ and the aggregated context encoding. While the CNP feeds the output of $a$ directly to $\rho$, the NP first maps it to a distribution from which the latent variable $z$ is sampled to be fed to the decoder. $\mu$ and $\sigma$ denote the means and variances of the respective distributions.}
\label{fig:background}
\end{figure*}

\textbf{Conditional Neural Processes (CNPs): } CNPs \cite{garnelo2018conditional}  model the  predictive distribution  by constructing a deterministic mapping from the context $C$ to $\theta$. Namely, $\theta$ encodes $C$ using the encoder composition $E = a \circ \phi$ where, $a$ and $\phi$ are the static average and the learnable neural network, respectively. The encoding $E(C) \in \mathbb{R}^e$ together with the locations of $T$ are  decoded by  $\rho$ to arrive at the predictions (fig. \ref{fig:cata}). CNPs thus  estimate the density parameters for $T$ using an encoder-decoder composition  $\rho \circ E$ with its  domain being the power set $2^{X_C}$:

        \begin{equation}
        \Phi(T) = \rho \left( E(C), T \right),
        \label{eq:equivdeepset}
        \end{equation}

where  $\Phi$ relies on deep neural networks as choices for $\rho$ and $\phi$ and is thus known as a Deep Set \cite{zaheer2017deep}. The set representation for context and target data further deems the valid choices of functions for the encoder $E$ (acting on the outputs of permutation-sensitive $\phi$) to be the ones whose outputs are indifferent to the ordering of the input elements. For a supervised learning problem setup, such functions can help predict $x_i \in T$ while remaining \textit{permutation invariant} w.r.t. the predictors. This property is a necessary condition for mimicking a stochastic process as it ensures \textit{exchangeability} in the predictive distribution $p(f(T) | T, C)$ (see App. \ref{app:stochastic_process} for the preconditions of stochastic processes).  Deep Sets thus define $E$ to be the symmetric average over the individual contextual embeddings $\phi_{i \in N}$:

        \begin{equation}
           E(C) = a(\phi(C)) =  \frac{1}{N} \sum_{(x,y) \in C} \phi_{xy}([x;y]),
           \label{eq:deepsetencoder}
        \end{equation}

where the encoder $\phi$ does away with the GP-styled analytical priors (see App \ref{app:gp} for a preliminary on GP) and instead extracts prior knowledge from the observations $[x;y]$ of $f$ empirically.
Different from other deep learning domains, the average pooling function $a$ here is crucial for capturing set-based representation and forms a special case of a larger family of pooling functions discussed later in Section \ref{sec:resemblances}. Given the permutation invariant context encoding $E(C)$, the predictive distribution can now be modeled by the decoder  as a factorized Gaussian across the target set  to satisfy the \textit{consistency}  of stochastic processes:
\begin{equation}
\begin{split}
    p&(f(T) | T, C) = \\ &\prod_{x \in T} \mathcal{N}(f(x) | \mu(\rho(x, E(C))),  \text{diag}[\sigma^2(\rho(x, E(C)))]),
    \end{split}
    \label{eq:factorized_gaussian}
\end{equation}
where $\mu(.)$ and $\sigma(.)$ denote the mean and the standard deviation of the multivariate Gaussian distribution with diagonal covariance matrix learned by the decoder. 

\begin{figure}[t!]
    \centering
      
        \includegraphics[width=0.25\textwidth]{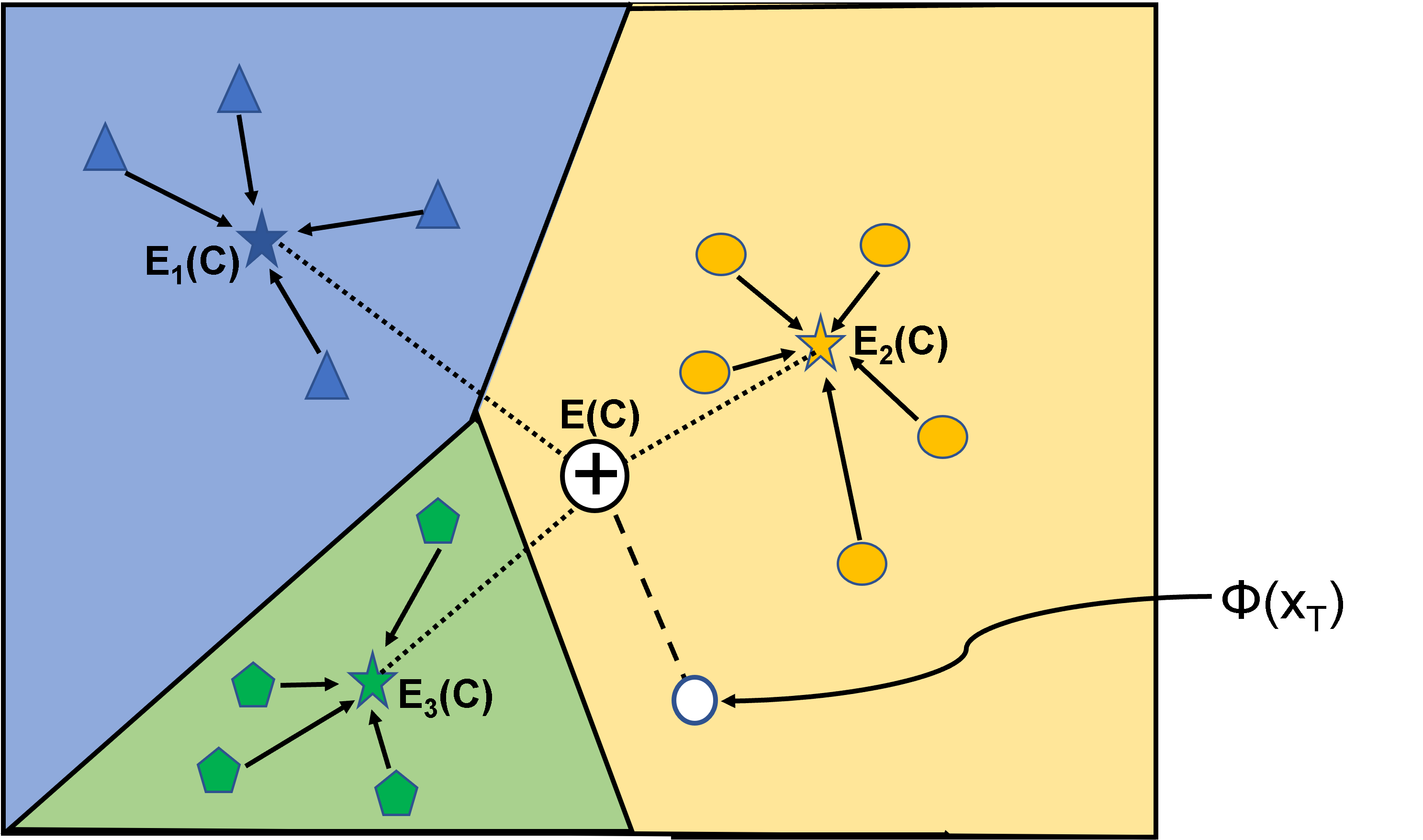}

\caption{CNP \cite{garnelo2018conditional} as a prototypical network \cite{snell2017prototypical}: the representations of a class $i$ are averaged  to form the class-specific prototype $E_i(C)$. The final context encoding $E(C)$ is a result of the concatenation of all  class prototypes $\oplus E_i(C)$ and is used to decode predictions for the query $x_T$. }
\label{fig:prototypical}%
\end{figure}

The use of the aggregated encoding for predictions  owes CNPs  a resemblance to the \textit{prototypical networks} \cite{snell2017prototypical}. CNPs  learn the metric space where, the context representations serve as  prototypes to match the queried target representation. In particular, the classification setup of CNPs \cite{garnelo2018conditional} groups the classwise context encodings   to form the support set of each class in the embedding space. The prototype  of each class is thus the mean of its support set (fig. \ref{fig:prototypical}). 

The optimization objective for the CNP involves finding a set of optimal parameters $\theta^*$  that maximizes the log predictive probability of observing targets given the  context:
\begin{equation}
    \theta^* = \underset{\theta}{\mathrm{arg\;max}}\, \mathbb{E}_{C, T, f(T)} [\log p(f_\theta(T) | T, E_\theta(C))],
    \label{eq:logprob}
\end{equation}
where $\mathbb{E}_{C, T, f(T)}$ is the expectation over distribution of datasets drawn from $D$, each of which contributes to the sampling of a unique $C$ and $T$.  Since this optimization objective is shared by all other NPF members, these meta learn the distribution of datasets in modeling the underlying stochastic process. Further, the inclusion of labels in the context $C$ owes the NPF members with generative modeling capabilities (see Section \ref{sec:resemblances} for the meta-learning and generative perspectives to the NPF). 

\textbf{(Latent) Neural Processes:}
 CNPs derive the fixed encoding $E(C)$ by processing each context point independently and then computing the predictive  distribution $p(f(T)|T, C)$), \textit{i.e.,} they perform \textit{point-wise} modeling. However, we might desire  to consider beyond point-wise modeling of inputs and instead exploit their correlation during prediction. For instance, in image inpainting, it is desirable to assign similar pixel intensities to nearby pixel locations. 
 
 There exist multiple ways to model the above correlation in the deep learning domain. One could  trade the parallelizability of predictions to instead generate predictions autoregressively \cite{van2016conditional, bruinsma2023autoregressive}. Alternately, 
one could draw an analogy to GPs, \textit{i.e.,} rather than learning the diagonal variances of the multivariate Gaussian  defining the input predictions, the model can be trained to learn the entire covariance among the input variables  \cite{yoo2021conditional}. Despite their powerful expressivity, these techniques scale poorly with data.\footnote{More recent NPF members enhance the tractability of autoregressive modeling by approximations of the full covariance matrix \cite{nguyen2022transformer} (see section \ref{sec:specific}).}
 
 To achieve correlation modeling, Garnelo \textit{et al.} \cite{garnelo2018conditional}  consider the rather intuitive possibility of a context set generating more than one function samples with different priors that can represent point-wise uncertainty on the target set equally well. This hints towards the fact that the priors can be governed by a yet another uncertainty and that modeling point-wise uncertainties might not be enough to determine them. Extending CNPs with latent Gaussian  distributions thus  help capture such \textit{global uncertainty} in the overall structure of the function (fig. \ref{fig:catb}). Samples from  the distributions  correspond to individual functions rather than points and are the enablers of diverse stochastic factors.  The resulting latent NPs incorporate a high-dimensional stochastic vector $z \sim \mathcal{Z}$ that captures all of the context information while inducing  randomness into the posterior of the functions.\footnote{We use the acronym NP interchangeably for the latent NP as well as for the generic NPF models.}   In other words, each individual $z$  results in a deterministic decoder function $\rho(x, z_i)$. This also offers us a Bayesian perspective to inspect $z$ since $p(z)$ now  encodes the data-specific prior learned by the model over the context. On observing the target, such a prior gets updated to form the posterior $p(z|C,T)$ that can better estimate the \textit{epistemic uncertainty} due to the lack of ample context data. 

$p(z|C, T)$ is, however, intractable in high-dimensional spaces due to the overtly complicated \textit{evidence} $p(C)$. A well-known solution to computing $p(z|C, T)$ remains approximating it with a variational posterior $q(z|C, T)$ of $z$. Such a posterior is parameterized by an encoder network $\phi$ that is permutation invariant over $C$. The variational prior $q(z)$ then initializes the parameters of $\phi$ to follow a well-behaved distribution, often a standard multivariate Gaussian. 
  $z$ can thus be modeled using the factorized Gaussian which, similar to eq. \eqref{eq:factorized_gaussian}, is a function of the  mean $\mu$ and variance $\sigma^2$ guaranteeing \textit{exchangeability} to the statistics $C$ of the context:

\begin{align}
\begin{split}
            p(f(T)|T, C) &= \int p(f(T) | T, z)p(z|C)dz \\
        &\approx \int p(f(T) | T, z)q(z|C)dz  \\
       \text{s.t., } q(z|C) &= \mathcal{N}(z|\mu(C), \text{diag}[\sigma^2(C)]),
\end{split}
\label{eq:posterior}
\end{align}
where $p(f(T) | T, z)$ is the posterior predictive likelihood of $z$ parameterized by a decoder  in a fashion similar to eq. \eqref{eq:factorized_gaussian}.  The generative model ensuring \textit{consistency} to $z$ becomes:
\begin{equation}
\begin{split}
    p(z, &f(T) | T, C) = \\&p(z|C) \prod_{x \in T}^{} \mathcal{N}(f(x)|\mu(\rho(x, z)), \text{diag}[\sigma^2(\rho(x, z))]),
    \end{split}
    \label{eq:generative_model}
\end{equation}
where $p(z|C) := \mathcal{N}(z|\mu_z, \sigma_z^2)$ is a multivariate standard normal capturing the global uncertainty of the functional distribution.
The optimization objective of an NP thus maximizes the following evidence lower bound of the log probability of the predictive distribution:  
\begin{equation}
\begin{split}
    \log \;&p(f(T) | T, C) \geq \\ 
    & \mathbb{E}_{q(z|C, T)} \left[ \sum_{x \in T}^{} \log p(f(x) | z, x) - \log \frac{q(z | C, T)}{ q(z | C)} \right], 
\end{split}
\label{eq:kldiv}
\end{equation}
where (C,T) is the observed dataset and $\log \frac{q(z | C, T)}{ q(z | C)}=D_{KL}$ is the reverse KL-divergence between the approximate conditional prior $q(z|C)$ and the approximate posterior $q(z|C,T)$. The approximate prior  replaces the true  prior $p(z|C)$ given the condition on the encoding $C$ implies a lack of exact prior.  In order to optimize eq. \eqref{eq:kldiv},  NPs backpropagate the  gradients using the reparameterization trick \cite{Kingma2014} on a single Monte Carlo sample of the approximated posterior.

\subsection{NPF relating to other deep learning domains}
\label{sec:resemblances}

\begin{figure}[t!]
    \centering
    \includegraphics[width=0.25\textwidth]{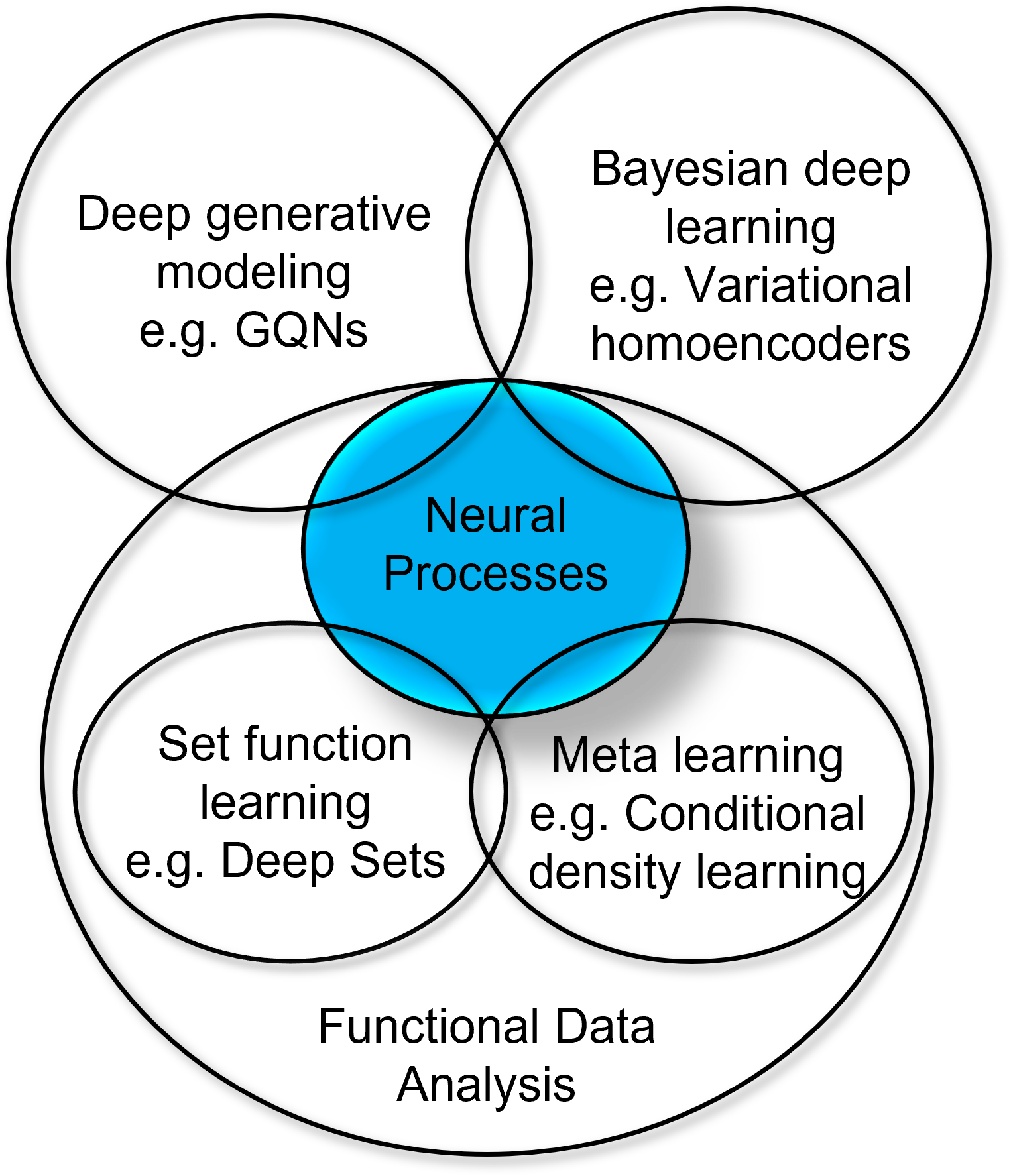}
    \caption{NPs viewed  from other deep learning domains resemble closely to: deep generative models like GQNS \cite{eslami2018neural}, Bayesian inference methods like variational homoencoders \cite{hewitt2018the}, set function approximators like Deep Sets \cite{zaheer2017deep}, and model-based conditional density meta learners \cite{kernelmean}.  }
    \label{fig:viewpoints}
\end{figure}

This section surveys the deep learning research directions that are brought together under the NPF umbrella (see fig. \ref{fig:viewpoints}). We  discuss the capabilities of NPs for set-based representation learning, meta-learning, functional data analysis, regression analysis, Bayesian learning, and  generative modeling. We further point out  specific works along each of these directions that the  NPs resemble the most with.

\begin{figure}[t!]
\centering
\subfloat[Exhaustive Janossy pooling (k=N)]{\includegraphics[width=2in]{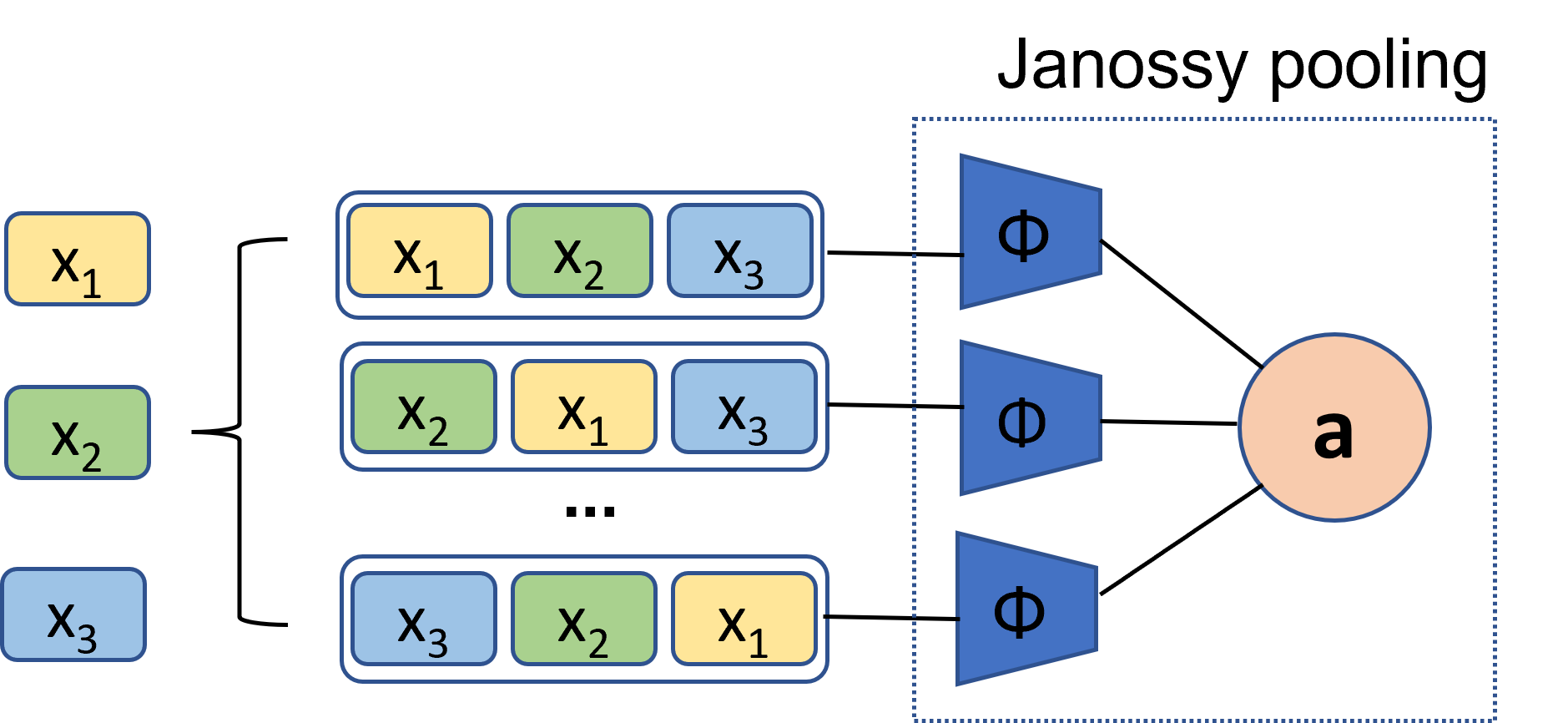}%
\label{fig:janossy_a}}
\\
\subfloat[NPs/CNPs (k=1)]{\includegraphics[width=1in]{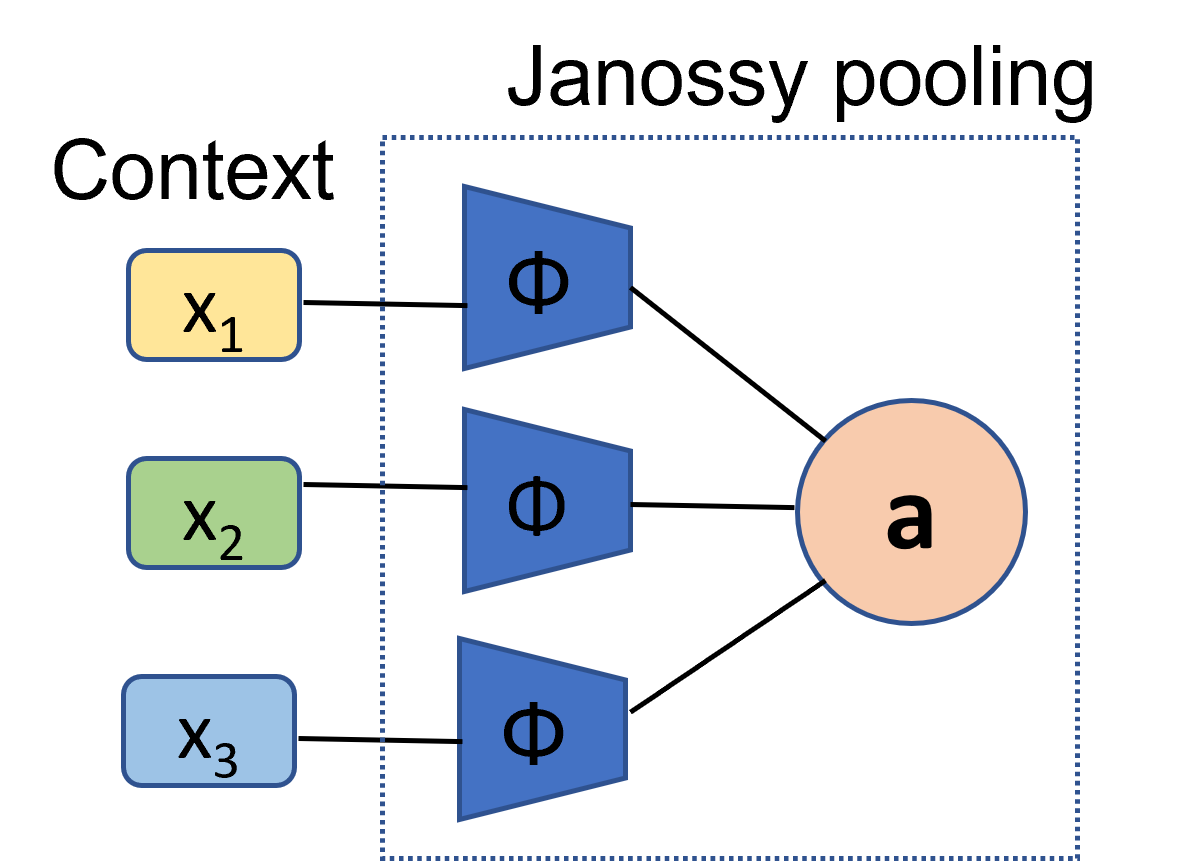}%
\label{fig:janossy_b}}
\hspace{0.2in}
\subfloat[ANPs (k=2)]{\includegraphics[width=1.7in]{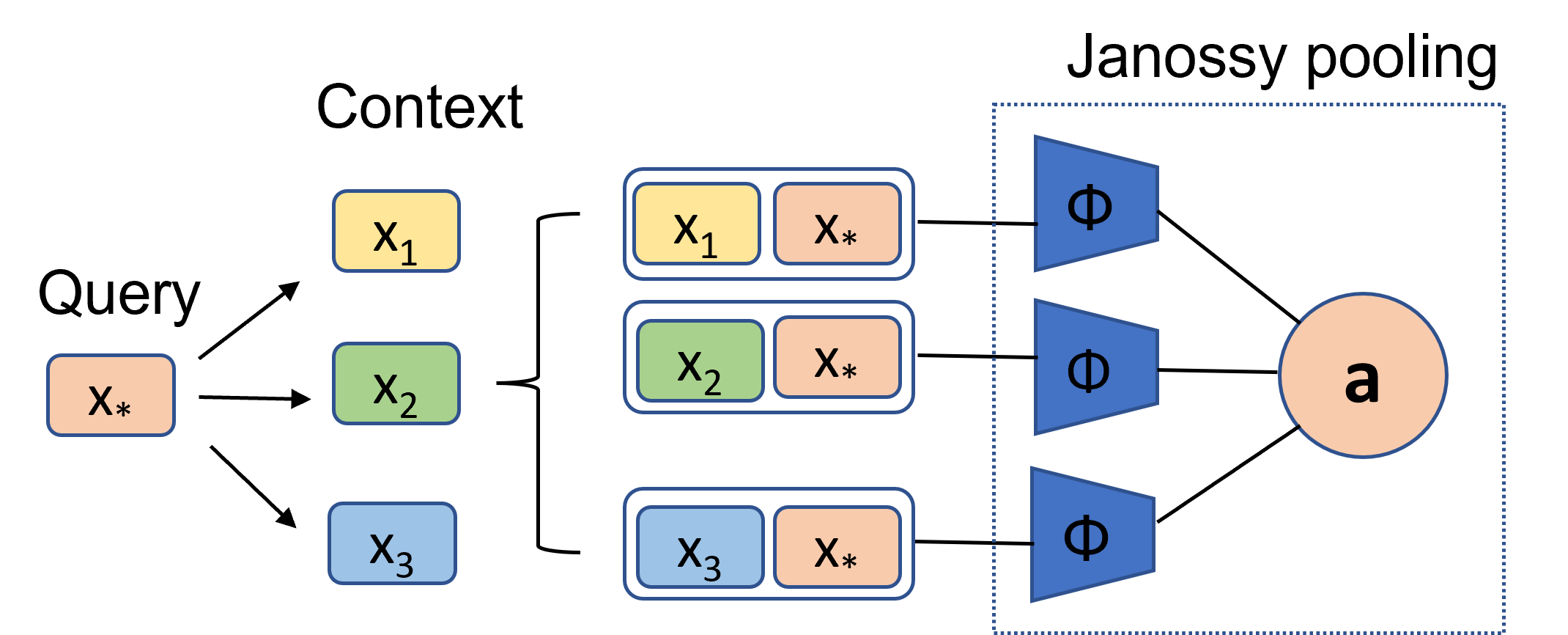}%
\label{fig:janossy_c}}
\caption{Janossy pooling and its instances for NP variants: $\phi$ is the permutation-sensitive encoder acting on inputs $\{x_i\}_{i =1}^N$ and $a$ is the averaging operation. The dotted boxes compose the pooling mechanism.}
\label{fig:janossy}
\end{figure}

\textbf{A. Set-based learning. } NPs operate on variable length context sets by encoding them using order-sensitive neural network parameters. The need for an additional permutation-invariant operation thus becomes imminent if the representation is to be independent of the input order. To address this, NPs exploit the Janossy pooling architecture
\cite{murphy2018janossy} that  derives a permutation-invariant approximation for an input set $\{x_1, ..., x_N\}$  by considering all  possible permutations $\pi(x_{1:N})$ of it, applying the permutation-sensitive encoding $\phi$  and averaging the encodings (see fig. \ref{fig:janossy_a}). To alleviate the $\mathcal{O}(N!)$  complexity of  the exhaustive reorderings of the sequence,  $k-$order  permutations are instead preferred. Deep Sets and self-attention thus form  $k$-ary Janossy pooling with the respective $k$ values being 1 and 2 \cite{wagstaff2021universal}. 
 CNPs \cite{garnelo2018conditional} and  NPs \cite{garnelo2018neural} feed each context point individually to $\phi$, \textit{i.e.,} $k=1$ (see fig. \ref{fig:janossy_b}). While this makes for the cheapest computational instance of Janossy pooling, using $k=1$ removes any possible inductive bias from the decoder that could have helped encoding interactions among the inputs. As a result, the decoder for the CNP and  NP must explicitly learn to reason relationship among the input samples during training.  However, such decoder architectures   fail to induce relational reasoning \cite{wagstaff2021universal}. 
Attentive Neural Processes (ANPs) \cite{kim2018attentive}, discussed in section \ref{sec:specific}, rely on self-attention mechanisms to perform input-dependent weighted pooling of information and thus induce a more implicit pairwise ($k=2$) relational reasoning of inputs into the decoder. This in turn allows them to map sets of points to sets of points \cite{lee2019set}.   Fig. \ref{fig:janossy_c} illustrates ANPs where the Janossy pooling is embodied by the cross-attention operation between the query and the self-attended context inputs.

\textbf{B. Meta-learning. }  NPs mimic the capabilities of GPs to capture distribution over functions using the powerful approximation capabilities of neural networks. To do so, they 
 \textit{meta-learn} set functions on several learning episodes of diverse tasks and can thus generalize well to new unseen tasks after seeing only a few examples of these. Each training task models an underlying stochastic data generator function and comprises learning from the labelled context set C to be able to make predictions on a set of unlabelled target set $T$.  Learning over multiple tasks comprising $C$ and $T$  helps them select inductive biases from data. For instance, training NPs for image completion with subsequent tasks consisting  of images of the same object(s) captured under different conditions such as size, angle, etc. can induce group symmetry to scaling and rotation \cite{kawano2021group}. In the standard meta-learning literature, $C$ and $T$ are often synonymous to the support and the query set, respectively.
 
 NPs  meta learn stochastic processes by  employing a \textit{model-based}  approach  \cite{Huisman2022}. However, viewing NPs in the light of prototypical networks (as noted in section \ref{sec:preliminaries}) does open up the possibility of \textit{metric-based} meta-learning.  A meta learner rather close to NPs are thus the conditional mean embedding operators learning density estimation \cite{kernelmean}.  The model-based meta learner perspective further helps explain the superior efficiency of NPs to gradient-based learners \cite{finn2017model, nichol2018reptile} that typically demand a significant amount of memory and computation besides suffering from model sensitivity \cite{antoniou2018how} and local optimum affinity \cite{molybog2021does}.

 \textbf{C. Functional data analysis. } NPs perform \textit{functional data analysis} (FDA) by meta learning distributions of (potentially discretized \cite{Gordon2020Convolutional}) data generating functions sampled irregularly along a continuum (time, space, wavelength spectra, etc.). Analyzing functional data however involves dealing with the non-linearity of the data dimensions coupled with their intricate inter-dimensional dependencies. Subsequently, FDA is known to be a notoriously complicated task for standard machine learning algorithms \cite{lin2019mfpca} and challenging even for deep neural networks \cite{rossi2005representation}. NPs thus stand out in their capacity to encode the samples of infinite dimensional functional data using the finite dimensional encodings $E(C)$ obtained from the  respective Janossy pooling instantiations.  NPF members therefore not only complement the recent advances in functional autoencoding \cite{hsieh2021functional, hsieh2021srvarm} but also open up the doors to achieve \textit{practical} universal approximation of set functions -- practical because of the pooled encoding is known to underfit the context set \cite{kim2018attentive} despite their theoretical plausibility to perform otherwise \cite{zaheer2017deep, bloem2020probabilistic}. It is the latter problem  that has in turn been the inspiration behind extending finite dimensional encodings (see eq. \eqref{eq:deepsetencoder}) to infinite-dimensional functional representations \cite{Gordon2020Convolutional, xu2020metafun}.

\textbf{D. Regression algorithms. } While NPs have been  applied to a number of FDA tasks including classification and clustering \cite{pakman20a}, they were primarily introduced as \textit{regression algorithms} \cite{garnelo2018conditional} that learn the  stochastic processes characterized by arbitrary polynomial functions $f$ using neural networks.
 As such, a number of other tasks can be regarded as specific cases of regression. For instance, classification requires a step function regression   where, the global aggregation operation in eq. \eqref{eq:deepsetencoder} is replaced by a classwise aggregation of encodings \cite{garnelo2018conditional}. 
 Using neural networks for regression endows NPs with two major benefits over the GPs: (a)  their data-driven approach to approximate fairly complicated functions and (b) their efficiency at inference.   NPs as regression algorithms further align well with  deep kernel learning \cite{wilson2016deep}.   In fact, affine-decoder NPs are mathematically equivalent to GPs with deep kernels  \cite{rudner2018connection, tsymbalov2019deeper}. 
 
 \textbf{E. Bayesian modeling. } The regression capability  offers a statistical \textit{Bayesian} window to NPs by  positing an empirical prior over the context.
 Further, the meta-learning setup  requires a hierarchy in the Bayesian model in order to infer parameters shared across tasks \cite{grant2018recasting}. For latent variable NPs, optimizing  the Bayesian hierarchy means approximating the intractable posterior $p_\theta(z|C,T)$ with a variational posterior $q_\phi(z|C,T)$ (see eq. \eqref{eq:kldiv}). This Bayesian training objective endows NPs with greater efficiency  over non-Bayesian  objectives  that often suffer from curse of dimensionality \cite{fong2020marginal}.  Further, the Bayesian perspective puts NPs side by side with popular variational inference models including partial variational autoencoders \cite{ma2019eddi},  variational homoencoders \cite{hewitt2018the} and variational implicit processes \cite{vips}.

\begin{figure}[t!]
    \centering
    \includegraphics[width=0.25\textwidth]{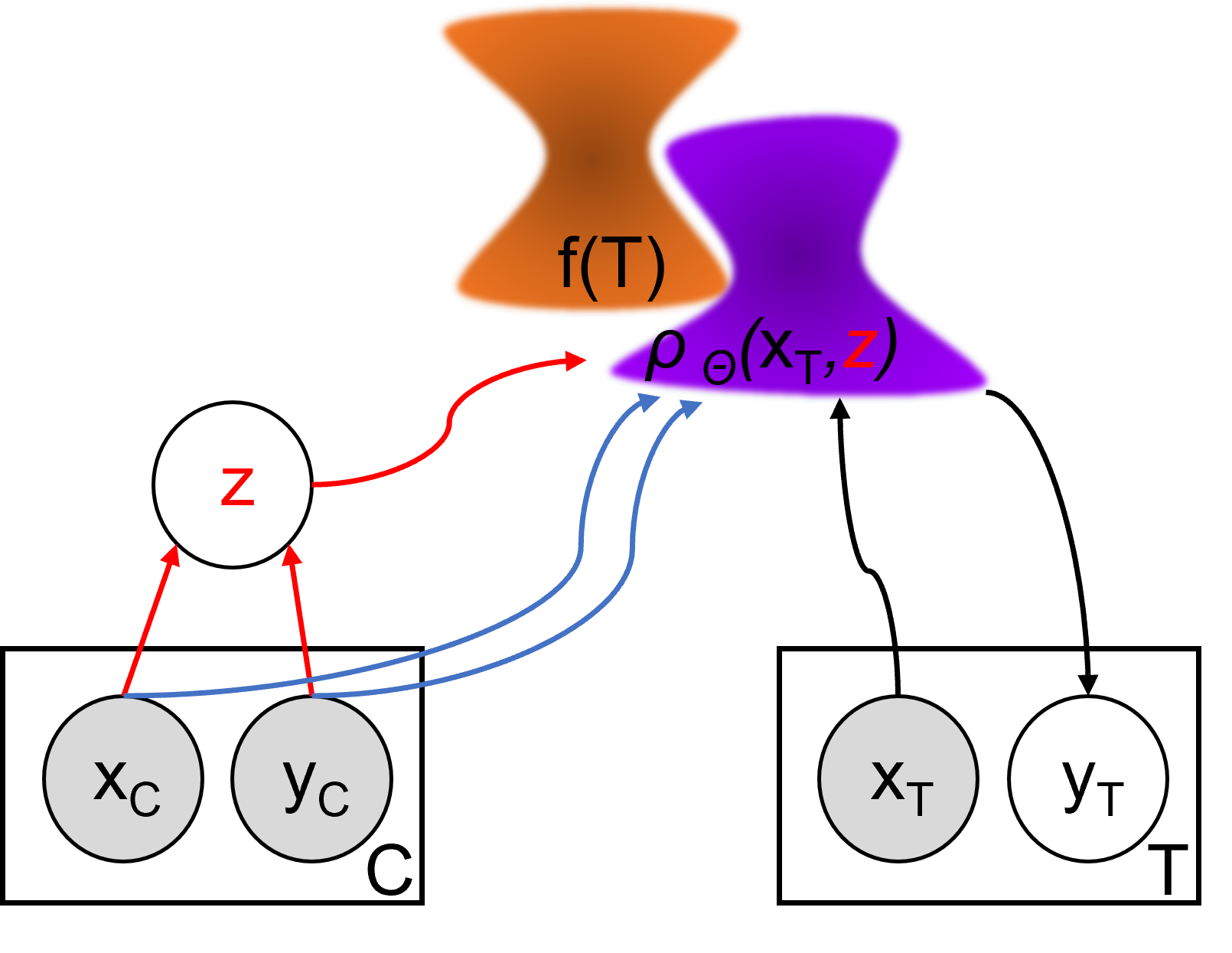}
    \caption{Graphical model depicting the CNP and latent NP as deep generative models. Gray shades stand for observed variables. Colors differentiate model-specific setting: \textcolor{red}{red} for latent NPs, \textcolor{blue}{blue} for CNPs, and black common to both. For latent NPs, $\rho_\theta$ is the decoder mapping the samples from the relatively simple distribution of $z$ to the intricate distribution $\rho_\theta(z)$. Deep generative models are trained with an objective to make $\rho_\theta(x_T, \textcolor{red}{z})$ resemble the true  distribution $p(f(x))$ (shown in \textcolor{orange}{orange}) by maximizing the likelihood $p(f(x)|z)$ of all observed $x \in C \cup T$.}
    \label{fig:deep_gen}
\end{figure}

\textbf{E. Generative modeling. } NPs  are likelihood-based \textit{deep generative models} that learn the intricate probability distribution of $f$ based on a finite number of independently and identically distributed (i.i.d.) observations $(X_{C}, Y_C) \sim D$. The NP modeling objective involves  mapping locations $x$ and their values $y$ from the tractable distribution of latent variables $z \sim \mathcal{N(\mu, \sigma^2)}$ supported in $\mathbb{R}^q$ to points in $\mathbb{R}^{d_y}$ that resemble the given labels $Y_C$. A latent NP is thus a generator $g$ that samples $z$ and computes $g(z)$ based on eq.  \eqref{eq:generative_model} \cite{ruthotto2021introduction}. This amounts to computing the evidence (eq. \eqref{eq:posterior}) of a sample $x \in T$ whose observation likelihood $p(f(x)|x, z)$ accounts for the similarity of $g(z)$ to $f(x_T)$ (see fig. \ref{fig:deep_gen}).  

Meta-learning generative properties enables NPs to learn the true data distribution from limited observations. This  endows them a resemblance with other minimal supervision frameworks including  neural statisticians \cite{edwards2016towards} and generative query networks (GQNs) \cite{eslami2018neural}. For instance, GQNs can be seen as specific cases of NPs that  model the distribution over functions generating scenes rather than those over the generic family of random functions. Subsequently,    such probabilistic generative models can be put under a common verification framework that imposes the requirement that for every choice of conditioning input to the model, their outputs must satisfy a linear constraint with high probability over the sampling of latent variables \cite{djprob}.
\section{Taxonomy of the Neural Process Family}
\label{sec:classification}
\begin{figure*}[h!]
    \centering
    \footnotesize
\begin{tikzpicture}[
  level 1/.style={sibling distance=21mm}, 
  edge from parent/.style={->,draw},
  >=latex, level distance = 12mm]
  
\node[root] {Inductive biases}
    child{ node[level 2] (c1) {\textbf{Sec. \ref{sec:agnostic}} Task agnostic encoding}}
    child {node[level 2] (c2) {\textbf{Sec. \ref{sec:specific}} Task specific encoding}}
    child{ node[level 2] (c3) {\textbf{Sec. \ref{sec:data_structure}} Symmetry equivariance}}
    child{ node[level 2] (c4) {\textbf{Sec. \ref{sec:relational_bias}} Graph connectivity}}
    child{ node[level 2] (c5) {\textbf{Sec. \ref{sec:pred_corr}} Output dependency}}
    child {node[level 2] (c6) {\textbf{Sec. \ref{sec:temp_correlation}} Temporal dependency}}
     child{ node[level 2] (c7) {\textbf{Sec. \ref{sec:task_relatedness}} Multi-task relatedness}}
    child{ node[level 2] (c8) {\textbf{Sec. \ref{sec:domain_invariance}} Domain Invariance}};
\begin{scope}[every node/.style={level 3}]
\node [below of = c1, yshift=-1pt] (c11) {Learn set-based representation \cite{garnelo2018conditional},   \cite{carr2019wasserstein}};
\node [below of = c2] (c21) {Reason relation among inputs \cite{kim2018attentive}, \cite{sureshimproved}, \cite{lee2020residual}, \cite{yoon2020robustifying}, \cite{wang2020doubly},  \cite{kim2022neural}, \cite{yu2022research}, \cite{nguyen2022transformer}};
\node [below of = c3] (c31) {Mimic input transformations in outputs \cite{Gordon2020Convolutional}, \cite{kawano2021group}, \cite{holderrieth2021equivariant}};
\node [below of = c4] (c41) {Learn connectivity among inputs' neighbors \cite{nassar2018conditional}, \cite{carr2019graph}, \cite{Louizos2019TheFN}, \cite{DBLP:journals/corr/abs-2009-13895},  \cite{liang2021neural}, \cite{yoo2021conditional}  };
\node [below of = c5] (c51) {Model input correlations \cite{garnelo2018neural},   \cite{lee2020bootstrapping}, \cite{foong2020meta}, \cite{wang2021np},  \cite{bruinsma2021the}, \cite{wang2021global}, \cite{markou2022practical}};
\node [below of = c6] (c61) {Model temporal sequences \cite{singh2019sequential}, \cite{willi2019recurrent}, \cite{DBLP:journals/corr/abs-1910-09323},  \cite{norcliffe2021neural}, \cite{pmlr-v161-petersen21a}};
 \node [below of = c7] (c71) {Learn fine-grained task representations \cite{requeima2019fast},  \cite{kim2022multitask}};
\node [below of = c8] (c81) {Learn generic domain representation \cite{kallidromitis2021contrastive}, \cite{gondal2021function}, 
 \cite{mathieu2021contrastive}, \cite{ye2022contrastive}, \cite{gao2022matters}};

\end{scope}

\foreach \value in {1,...,8}
  \draw[->] (c\value) -> (c\value1);

\end{tikzpicture}
\caption{\hspace{0.2in}A taxonomy of the members of the Neural Process Family on the basis of the primary inductive biases they target.}
\label{fig:classification}
\end{figure*}
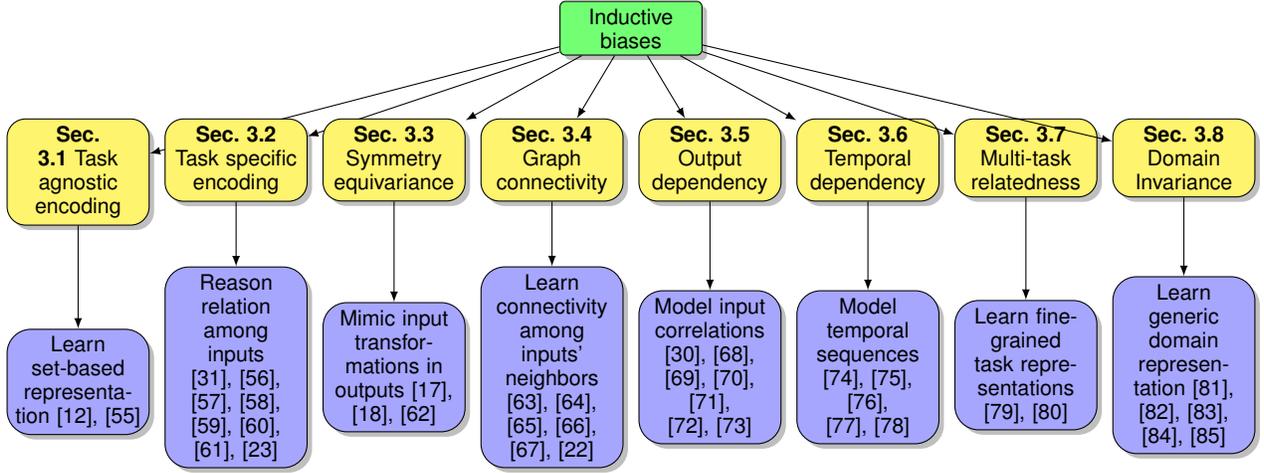

We now present a detailed survey of the  advances in the Neural Process Family (NPF).
Given the broad scope of these works, we first classify these by exploiting the fact that all  NPF members encode a set of assumptions as inductive biases to generalize the finite context set into a model suitable for the input domain. For example, CNPs \cite{garnelo2018conditional} use  the averaged encodings to capture set  representation while the latent NPs \cite{garnelo2018neural} additionally maintain a global input  distribution   to induce the correlation among its predictions. By the same token, we identify a subtotal of eight major inductive biases encompassing the NPF tree (fig. \ref{fig:classification}).

\subsection{Task agnostic set function approximation}
\label{sec:agnostic}

This category consists of the generic methods using Deep Sets \cite{zaheer2017deep} to model complex functions. The groundbreaking CNPs \cite{garnelo2018conditional} (see Section \ref{sec:background}) that  models the parameterized distribution over functions using a fixed real-value vector falls into this category.  CNPs assume the distribution domains to be overlapping and hence, minimize the reverse KL divergence between the true and the approximate posterior during training. The mode-seeking behaviour of reverse KL helps optimize the blown up loss values (due to $\log(q(z|E(C))) = 0$ in eq. \eqref{eq:kldiv}). 

However, there could be scenarios where, a poorly specified model leads to a zero or an intractable data likelihood thus deeming the domains of the modeled and the exact distribution to be disjoint. This could drive both forward and reverse KL values to infinity. While a naive getaway   could be using the Jensen-Shannon (JS) divergence as a proxy of KL to cap the loss values to $\log(2)$, the saturation of JS losses renders the gradients unusable \cite{kolouri2018sliced}.  Recent  works thus improve the posterior approximation objectives by either opting for a surrogate objective of the target log-likelihood  \cite{wang2023bridge} or by avoiding the need for the likelihood measurement altogether. Along the latter path,     Wasserstein NPs (WNPs) \cite{carr2019wasserstein}   minimize the sliced Wasserstein distance \cite{7780937} between the decoded  and the actual set of predictions instead of maximizing eq. \eqref{eq:kldiv}. 
 
  Finally, it is worth acknowledging that there is a line of work on NPF models which looks at different objective functions rather than just different inductive biases. The WNPs \cite{carr2019wasserstein} are one  example of these. Similarly, the ConvNP \cite{foong2020meta} discussed in section \ref{sec:pred_corr} gets away with variational posterior inference and instead approximates the biased log likelihoods of observations using importance sampling.

\subsection{Task specific set function approximation}
\label{sec:specific}
\par
While the  task-agnostic representation allows NPs to encode a set function $f$, these provide little clue on the context-query relationship, \textit{i.e.,} the methods relying on task-agnostic set encoding  assign equal weight to all the observations while predicting the value of a given query. This involves processing each context point individually to produce a finite   $\mathbb{R}^e$-dimensional context encoding $\phi_{i \in N}$. Kim \textit{et al.} \cite{kim2018attentive} show that the resulting encodings underfit the observed context set as the individual processing of data points takes away the capability of the model to perform relational reasoning.  Furthermore,  the cardinality $N$ of the context set forms a theoretical lower bound for $\mathbb{R}^{e}$ to universally represent an arbitrary $f$ even in the face of encoders and decoders with unlimited flexibility \cite{wagstaff2019limitations}. Given the practical limits to  flexibility, even $\mathbb{R}^e > N$ can not guarantee task-agnostic function approximators to  represent all $f$.

Task-specific set function approximators overcome the aforementioned limitations by enriching the NP encodings to be sensitive to the queried locations. A prime example of this category is the attentive NP \cite{kim2018attentive}  that uses scaled dot-product attention \cite{vaswani2017attention} as an inductive bias to model the pairwise relation among the inputs. ANPs use two steps to ensure that the encoding $E(x|C)$ is locally relevant per target point $x \in T$ rather than being globally static across the target set. First, a similarity is computed between the context input embeddings forming the keys $K$, and the target (in case of cross-attention) or the context (in case of self-attention) input embeddings forming the queries $Q$. The input embeddings then form the values $V$  to predict the local target embedding. ANPs further bind the potential of CNPs and latent NPs by leveraging the multi-head attention mechanism along the deterministic and the latent path.  The resulting predictive distribution can be given as: 

\begin{equation}
    \begin{split}
        p(f(T)| T, C) \approx \int p(f(T) | T, z, r_C^*) q(z | E_{SA}(C))dz, \\
        p(f(T) | T, z, r_C^*) = \prod_{x \in T} \mathcal{N}(y_t | \mu(x, r_C^*, z), \text{diag}[\sigma(x, r_C^*, z)]^2), \\
        q(z|E_{SA}(C)) = \mathcal{N}(z|\mu(E_{SA}(C)), \text{diag}[\sigma(E_{SA}(C))]^2),
    \end{split}
    \label{eq:attentivenp}
\end{equation}
where $E_{SA}(C)$ is computed in a way similar to eq. \eqref{eq:deepsetencoder} except for the operation $a$  now  replaced by trainable self-attention (SA) networks. To enhance the expressivity of  eq. \eqref{eq:attentivenp}, Suresh \textit{et al.} \cite{sureshimproved} propose the ANP\textsuperscript{++} model that adapts the decoded targets’ internal dependencies with respect to each other by using SA-based decoder.   

However, ANPs scale poorly with high dimensional data as the complexity of dot product attention evaluates to $\mathcal{O}(N(N+M))$ due to the multiplication $QK^T$ across the context-context and the  target-context pairs. In contrast, CNPs and NPs scale linearly with data. This makes using ANPs infeasible on higher dimensional tasks such as 2-d image completion \cite{Gordon2020Convolutional}.\footnote{Using global attention on higher resolution images has been a long-standing  issue in the domain of computer vision. For instance, a $256 \times 256$ ImageNet \cite{deng2009imagenet} input can involve computing global attention weights in the order of $256^4$, \textit{i.e.,} each of the $256 \times 256$ pixels attending to all the $256 \times 256$ pixels.}
 Inspired by the success of Vision Transformers (ViTs) \cite{dosovitskiy2021an} on images, Yu \textit{et al.} \cite{yu2022research} \textit{sketch} splitting an image into equal-sized context $P_C$ and target $P_T$ patches to compute patch representation vectors. The patch vectors along with their positional encoding are to be employed in the latent and deterministic encoder paths to derive the respective input summaries. Similarly, the decoder is to rely on the summaries alongside the positional encoding  of the target patch to compute predictions. \footnote{
At the time of writing, Yu \textit{et al.} \cite{yu2022research}  merely sketch the above architecture without any empirical validation of its performance. }

 ANPs, albeit to a lesser extent than NPs, are still prone to  underfitting  the context. Lee \textit{et al.}
 \cite{lee2020residual} show that the non-linear self-attention in ANPs  fail to approximate  the optimal summary $E_{SA}^*(C)$ of an arbitrarily linear system.  The authors propose Residual Neural Processes (RNPs) that exploit the structural resemblance of ANPs to Bayesian Last Layers (BLLs) \cite{weber2018optimizing}. 
Given the heavy reliance of BLLs on the context for computing posterior probability of observing a class  $c$, these have restricted modeling potential compared to ANPs. RNPs bring the best of both ANPs and BLLs  by first employing ANPs to predict the non-linear residual of $E_{SA}(t|C)$ and then using BLLs to derive the exact  $E_{SA}(C)$. 

Further on underfitting,    the noise sensitivity of ANPs can be seen to interfere with capturing of contextual embeddings for noisy data and restrictive task distributions \cite{kim2021neural, kim2022neural}. One way to resolve this is by replacing dot-product attention with stochastic attention  \cite{fan2020bayesian} in the ANP encoder. 
More recent methods such as the Transformer NPs (TNPs) \cite{nguyen2022transformer} address this by trading computational tractability for expressivity, and considering all input points $\{x_i, y_i\}_{i=1}^{N+M}$ as an ordered sequence. TNPs thus model the target points autoregressively by replacing the  latent variables altogether with fully attentive Transformer \cite{vaswani2017attention} architectures. 

Over the years, task-specific function approximating NPs have gained popularity across a number of real-world problems. For sequential modeling, Yoon \textit{et al.} \cite{yoon2020robustifying} show that the NP-based Sequential Neural Processes (SNPs) \cite{singh2019sequential} (introduced in section \ref{sec:temp_correlation})  are vulnerable to underfitting. As a result, the authors robustify SNPs by generating imaginary context that serve as proxies for applying attention \cite{rumbert1986learning}. For multi-task learning, several methods consider employing task-specific latents in addition to a global latent for modeling fine-grained intra-task stochastic factors \cite{wang2020doubly, kim2022multitask}.

\subsection{Euclidean symmetry equivariance}
\label{sec:data_structure}

This category exploits the symmetry of the data lying on Euclidean spaces to ensure that the predictions are equivariant to the input transformations. 
  Based on the type of equivariance targeted, we further divide this category of the NPF into two major sub-branches: the specific shift equivariant and the more generic group equivariant NPs.

\textbf{Shift-equivariant NPs.} This sub-branch aims to introduce shift equivariant convolutional neural networks (CNNs) into the NPF architectures given their huge success behind modeling visual signals \cite{lecun1998gradient, McGreivy2022ConvolutionalLA}. However, using CNNs  within NPF architectures is not straightforward. Given that NPs learn a function $f: \mathcal{X} \rightarrow \mathcal{Y}$ over context $C$, mirroring a shift on $f$ is effective only if the contextual summary $E(C)$ were to reside on a functional space (rather than a vector space). To address this, Gordon \textit{et al.} \cite{Gordon2020Convolutional} extend Deep Sets $\Phi$ to Convolutional Deep Sets (ConvDeepSets) $\Phi_{conv}$ that produce functional input embeddings. Unlike MLP-based $\Phi$ operating on continuous input-output spaces, $\Phi_{conv}$ uses CNNs for discrete  input-output mapping. The discreteness of CNN filters however implies varying overheads in handling a range of modalities, \textit{e.g.},   continuous  signals like images can be easily discretized  by reckoning the pixel locations while for  a  continuous time series signal,   discretizing involves the overhead of mapping the (discrete) CNN outputs back to a continuous  space.

    Formally, if $\hat{x} \in \hat{X}_C$ are the context locations obtained by discretizing the input space $x \in X_C$, then the ConvDeepSet $\Phi_{conv}$ bears the architecture of eq. \eqref{eq:equivdeepset} except for the encoding $E(C)$ lying  on the infinite-dimensional functional space: 
                \begin{equation}
            E(C)(\hat{x}) = \sum_{(x,y) \in C} \phi_y(y)\psi(x-\hat{x}),
        \label{eq:convdeepset}
        \end{equation}
    where $\phi.\psi$ makes up the encoding operation. 
    From the perspective of task-specific function approximation, the encoder $\phi$  can be viewed as an attention network that projects the label $y_i$ of a context sample on to a discrete space $\mathcal{S}$ based on the similarity between a discretized  query $x_\mathcal{S}$ and a context key $x \in X_C$. The resulting deterministic ConvDeepSet representation $\Phi_{conv}^\mathcal{S}$ is then passed through the filter $\psi$. Similarly, the decoder constitutes another ConvDeepSet whose discretized keys and values comprise $x_\mathcal{S}$ and $\Phi_{conv}^\mathcal{S}$ while queries being $x \in X_T$. ConvCNPs  however share the limitation of CNPs \cite{garnelo2018conditional} arising from the deterministic ConvDeepSet representation failing to capture diverse functional priors. To address this, Wang \textit{et al.} \cite{wang2021global} propose the Global Convolutional NPs (GB-CoNPs) that replace the deterministic projection space of  ConvDeepSets  with a latent space.

    \textbf{Group equivariant NPs.} This sub-branch addresses the limitation of CNN layers that are known to waste their precious weight-sharing capacity in learning redundant symmetries of the same feature templates \cite{olah2020naturally}.  Going beyond shift and inducing equivariances to a group of symmetries (e.g., rotation, scale, reflection, hue, etc.) can thus help optimize  on the network capacity of CNNs all while providing geometric guarantees that the learned representations remain stable to the groups of local and global transformations. To this end, Kawano \textit{et al.} \cite{kawano2021group} extend ConvCNP to EquivCNP by including further symmetries in scalar fields. EquivCNP  targets equivariance over the classical Lie Group $G$ by leveraging LieConv \cite{finzi2020generalizing} as the choice of convolutional layer.\footnote{ Lie group $G$ is preferred mainly for its smooth manifold with the mapping $L_g: G \rightarrow g, g \in G$  defined by $x \rightarrow gx$ \cite{finzi2020generalizing}.} In doing so, they  derive a continuous and group-equivariant extension of DeepSet, \textit{i.e.}, EquivDeepSet $\Phi_{eq}$. While DeepSets \cite{zaheer2017deep} and ConvDeepSets \cite{Gordon2020Convolutional} operate on vector and functional spaces, respectively, $\Phi_{eq}$  acts upon the homogenous space of $G$.

    Holderrieth \textit{et al.} \cite{holderrieth2021equivariant} propose Steerable CNPs (SteerCNPs) that further extend the scope of equivariance in the NPF to stochastic vector-valued fields. SteerCNPs  induce equivariance in the predictive posterior defined over the observed points in a stochastic field by exploiting the fact  that the prior over such a field is inherently invariant.   
    Similar to EquivCNP \cite{kawano2021group}, SteerCNPs leverage EquivDeepSet  and a $G-$equivariant decoder  over fiber input-output pairs. However, the choice of decoder is now a steerable CNN \cite{cohen2019general} and the target group of equivariances is now limited to rotations and reflections instead of the entire $G$ symmetries.
    
\subsection{Non-Euclidean graph connectivity}
\label{sec:relational_bias}
The NPF members produce functional encodings without taking neighborhood connectivity into account. For instance,  NPs aggregate independently processed encoding of the context points while ANPs capture relational reasoning among the points through fully-connected relational graphs of all possible pairs. These encodings fail to exploit the graph connectivity inductive bias  ingrained in the topology of a number of real-world problems including but not limited to traffic networks \cite{malecki2017graph},  social networks \cite{hunt2011using}, and cell dynamics \cite{bock2010generalized}. This NPF category thus includes models that explicitly induce non-linear connectivity biases of datasets. Given the natural tendency of graphs to represent such  connections, these make a popular choice of data structures for the methods in this category.\footnote{When viewed together,  symmetry equivariance (section \ref{sec:data_structure}) and  graph connectivity (section \ref{sec:relational_bias}) inducing biases can be seen to exploit the geometries underlying the data. These could thus be combined under the umbrella of  a geometry-inducing bias branch of the NPF taxonomy. However, since geometric deep learning usually concerns the non-Euclidean domains of graphs and manifolds, we consider listing these separately.}

Among the early works in this category include
Nassar \textit{et al.} \cite{nassar2018conditional} incorporating Graph Convolutional Networks (GCNs) \cite{Kipf2017SemiSupervisedCW} as encoders in the NP architecture to improve the learning of functions representing classes of graphs. Similar to CNPs, the proposed conditional graph NPs (CGNPs) feed on $N$ context and $M$ target points. However,  the context $C$ and target $T$ sets now also include their respective $N-n$ and $m-M$ neighborhood  points: $C = \{(v_i, y_i)\}_{i=1}^{n\geq N}; T = \{v_i\}_{i=n+1}^{n+m}$ where, $v_i$ and $y_i$ are the node attributes and labels. While processing a node $v_i \in C \cup T$, CGNP  encodes its $k-$hop neighborhood using message-passing isotropic bipartite GCNs \cite{Nassar2018HierarchicalBG}.  CGNPs thus generalize CNPs with the latter operating on    edgeless graphs where each node  connects only to itself. 

Similarly, Louizos \textit{et al.} \cite{Louizos2019TheFN} propose the Functional Neural Processes (FNPs)  that:  (a) adopt priors over the relational structure of the local latent variables instead of explicitly specifying a
prior  over (NP-styled)  global latents, (b) assume the inputs to lie on regular lattices and instead build a graph of dependencies among the latents that encode the input points. While (a) helps FNPs model dependencies among the input points by well-preserving the exchangeability (eq. \eqref{eq:deepsetencoder}) and consistency (eq. \eqref{eq:factorized_gaussian}) conditions, (b) makes FNPs resemble the autoencoder-based generative models \cite{Kingma2014}.\footnote{ANPs \cite{kim2018attentive} (see section \ref{sec:specific}) can  thus be seen to be a special case of the FNP with  deterministic versions of the context-context  graph \textbf{G} and the context-target graph \textbf{A} \cite{tang2021graphbased}. In this case, \textbf{G} models the self-attention among the context set elements while \textbf{A} models the cross-attention between the target and the context set elements.}

Graph structured data however comes with its own set of issues.
One such widely known issue  remains missing edge variable information \cite{burt1987note}. In favor of NPs,  replacing the missing values with conditional distribution samples can help enrich the neighborhood information \cite{huisman2009imputation}. To this end, Carr \textit{et al.} \cite{carr2019graph} employ NPs to impute the value distribution on edges by using the local structural eigenfeatures alongside the node attributes to describe an edge. 

Another prevalent issue with graphs is the problem of semi-supervised node classification in the face of limited labeled nodes. Day \textit{et al.}  \cite{DBLP:journals/corr/abs-2009-13895} address this by proposing the Message-Passing Neural Processes (MPNPs). MPNPs consider the context  to be $N$ labeled nodes and their $k-$hop neighbors while the target  includes all $N+M$ samples and their unlabeled neighbors, \textit{i.e.,} $T = \{v_i\}_{i=1}^{n+m}$. The latent samples $z$ in MPNPs  include the neighborhood information from the context encoding and thus capture the relational structure underlying the stochastic processes that generate the input. 
Similarly, Liang \textit{et al.} \cite{liang2021neural} propose NPs for Graph Neural Networks (NPGNNs) to predict the unknown links in the entire graph given the information about all the nodes (transductive) or only a subset of the latter (inductive).  NPGNNs combine CGNPs with MPNPs. Like CGNPs, NPGNNs employ message-passing GCNs \cite{Kipf2017SemiSupervisedCW} in their encoder. Like MPNPs, the context set encodings are aggregated to parameterize a global latent. The predictive probabilities of NPGNNs  thus describe the existence of an edge between the nodes $v_i$ and $v_j$.

\subsection{Output dependency}
\label{sec:pred_corr}

This category exploits inductive biases  that ensure the input correlation  is reflected in the model's predictions. Output dependency makes a great deal for real-world tasks like modeling of climate and brain signals  that demand capturing dependencies among a number of input variables over temporal and spatial dimensions. The latent NPs \cite{garnelo2018neural}  are the earliest such NPF members using a  latent variable to model the global input correlation. Inspired by these, Foong \textit{et al.} \cite{foong2020meta} propose  ConvNPs that extend  the ConvCNPs \cite{Gordon2020Convolutional}  with a latent path to additionally model shift equivariance. 

One limitation of latent variable NPs \cite{garnelo2018neural, kim2018attentive, singh2019sequential} is that they decode the predictive distribution relying on a single latent variable to capture the correlation among the function's positions $\mathcal{X}$ as well as its values $\mathcal{Y}$. Therefore, any distribution shifts in the random functions modeling future target sets can greatly increase the model's uncertainty. To remedy this, Wang \textit{et al.} \cite{wang2021np} propose handling the out-of-distribution  (OOD) data by decoupling the inference of the predictive distribution's variance from the function values and thus deriving the predictive means and variances from two distinct latent spaces, \textit{i.e.,} the decoder means are conditioned on the context values $Y_C$ as well as on the self/cross-correlations among the context-context and context-target locations, respectively while its variances are conditioned only on the latter cross correlations. 
Similarly, Lee \textit{et al.} \cite{lee2020bootstrapping} extend NPs with classic bootstrapping \cite{efron1992bootstrap} that models uncertainty by sampling a population with replacement  for the simulation of its distribution.  The resulting Bootstrapping Neural Processes (BNPs)  model functional uncertainty of a  dataset by bagging of the predictive uncertainties obtained using the bootstrapped contexts as inputs. 

Moreover, NPs are  liable to the intractable likelihood and complicated variational inference-based learning objectives arising from latent variable sampling. To bypass these intricacies,  Bruinsma \textit{et al.} \cite{bruinsma2021the} propose the subfamily of Gaussian NPs that replace latent variables with ConvCNPs \cite{Gordon2020Convolutional} to achieve predictive coherence.  Gaussian NPs  are capable of: (a) inducing shift equivariance in modeling the predictive distributions with Gaussian processes \cite{williams2006gaussian}, (b) hosting a closed-form likelihood while modeling correlations in the predictive distribution, (c) ensuring universal approximation like ConvCNPs, and (d) interpreting the standard maximum likelihood objective  as a well-behaved relaxation of the KL-divergence between stochastic processes \cite{matthews2016sparse}. 

On the lattermost capability, NPs can be seen to meta learn a Gaussian approximation $\Tilde{\pi}: \mathcal{O} \rightarrow \mathcal{D}_G$ of the \textit{posterior prediction map} $\pi_{f}: \mathcal{O} \rightarrow \mathcal{D}$ with $\mathcal{O}$ being the collection of observed datasets $O_{i=1}^{N+M} = \{(C, T)\}_{i=1}^{N+M} \sim f$ and $\mathcal{D}$ being the family of (global G) posteriors over  $f$. CNPs, on the other hand, learn  $\Tilde{\pi}$ by assuming the posteriors $\mathcal{D}_G$ to be obtained from deterministic priors and thus fail to model the predictive correlations. In contrast, Gaussian NPs allow more flexibility in modeling $\mathcal{D}_G$ by approximating $\Tilde{\pi}$ using a minimizer $\Tilde{\pi}(O)$ for the KL-divergence $D_{KL}$ between the predictive map $\pi_{f}$ and a Gaussian process $\mu$ for every $O \in \mathcal{O}$. Formally, to deal with the \textit{existence} of such minimizers for  high dimensional observations (where $D_{KL}$ is unbounded for the values around $\Tilde{\pi}(O)$), Gaussian NPs consider only finite-dimensional indices $|x|$ of $f$ to ensure the existence of an $|x|$-dimensional Gaussian distribution $\mu^x_G$ such that $D_{KL}(P_x\pi_{f}(O), \mu_G^x) < \infty$
where, $P_xf = (f(x_1), ..., f(x_{|x|}))$ projects $f$ onto the index set $x$. To  ensure that  $D_{KL}(\pi_{f}(D), \mu)$  approximates the KL minimizer,   the authors propose the basic Gaussian NPF member, a.k.a. the FullConvGNP that uses a CNN  as a positive semi-definite kernel $\mathbf{K}$. However, the  costly $2*d$ convolution operations limits the practical applications of FullConvGNPs to mostly 1-d data. Later variants overcome this  by  letting $\mathbf{K}$ to be convolution-free \cite{markou2021efficient, markou2022practical}.

\subsection{Temporal dependency}
\label{sec:temp_correlation}
This category induces NPs with biases to capture the temporal dynamics underlying sequences of stochastic processes, \textit{e.g.,} time-series modeling,  dynamic 3D scene rendering with objects interacting over time, etc. Formally, learning temporal dependency  demands modeling the set  $C_t = \{(x_i^t, y_i^t)\}_{i=1}^{N(C_t)}$  of observed context points drawn from the true stochastic process $f_t$ at a time step $t \in [1, S]$, \textit{i.e.,} $C$
now reflects the context and target sets over all time steps $t \in [1,S]$.
Furthermore,  the context size  is now variable over time and can even be zero for a given $t$ as such. Based on the nature of the induced temporal correlations, we subcategorize this branch into  implicit and explicit.

\textbf{Implicit time modeling. }
Singh \textit{et al.} \cite{singh2019sequential} propose Sequential Neural Processes (SNPs) that employ the Recurrent State-Space Model (RSSM) \cite{hafner2019learning} to capture temporal correlation across sequential context $C_1,..,C_S$ by transferring the knowledge from a task's context across temporal transitions of latent state-spaces. Like latent NPs, SNPs model $f_t$ as the standard distribution over latent variables $z_t$ conditioned on the given context set $C_t$. However, they additionally perform autoregressive modeling by conditioning the distribution over the latents of the past stochastic processes $z_{<t}$. This makes SNPs  to be a \textit{generalization} of NPs when considering either a unitary set of sequences,  or multiple sequences with no past context ($C_{<t} =$ {\fontfamily{qcr}\selectfont null}).

 While SNPs learn stochastic processes responsive to time scales, they are limited at modeling most real-world phenomena  incorporating hierarchies of time scales. For example, the electricity usage patterns for households  daily and across seasons. When modeled using a latent variable, such patterns lead to faster (daily) latents  being superimposed by slower (seasonal) latents.  Willi \textit{et al.} \cite{willi2019recurrent} 
 show that the addition of a recurrent hidden state per time step in SNPs not only lacks a principled way of propagating  temporally varying uncertainty but also restricts their expressivity by rather treating the modeling of temporal dynamics and stochastic process as two different tasks.
 Recurrent Neural Processes (RNPs) \cite{willi2019recurrent} address these by   extending SNPs with hierarchies of  state space models that can learn multiple layers of stacked stochastic processes by explicitly devoting a  latent variable to model the temporal dynamics among all of the spatial stochastic processes.

Similarly, Qin \textit{et al.} \cite{DBLP:journals/corr/abs-1910-09323} propose incorporating a parallel RNN path to the ANP \cite{kim2018attentive} to  capture uncertainty in predictions of temporally changing stochastic processes. The resulting ANP-RNN model uses the ANP to construct multiple observations of a stochastic process and is thus capable of uncovering the order underlying a sequential data. Further, the RNN path  leverages these observations to learn the uncertainty propagating through temporally progressive stochastic process.


\textbf{Explicit time modeling. }
 While the state-space models help capture  dynamically changing processes over time, such processes in real-world  systems are often governed by the infinitesimal changes in a number of intricate physical variables. It is therefore desirable to view the  optimization of the training objective in terms of infinitesimal  changes in parameters of the network. Modeling such changes in turn amounts to solving ordinary differential equations (ODEs) for the predictions of the network. Inspired by this, neural ODEs \cite{chen2018neural} rely on parameterizing the derivatives of ODEs  using neural networks. Neural ODEs thus offer an explicit treatment of time, a property lacking from the NPF literature in that NPs treat time as an unordered set and thus fail to measure the time-delay between real-world observations unless otherwise forced by the training objective. 
 
 However,   neural ODEs are  limited in their capacity to estimate uncertainties and to adapt quickly to the observation changes -- two properties inherent to NPs. To address this, Norcliffe \textit{et al.} \cite{norcliffe2021neural}  propose Neural Ordinary Differential Equations Processes (NDPs) that combine the benefits of NPs and Neural ODEs by leveraging NPs to parameterize the ODE derivatives. Namely, NDPs model the time-delays between different observations in low and high-dimensional time series by maintaining an adaptive data-dependent distribution over the underlying ODE. The NDP problem definition thus consists of modeling samples drawn irregularly from an underlying family of random functions $f: \mathcal{S} \rightarrow \mathcal{Y}$ where, $S = [t_0, \infty]$ and $\mathcal{Y} \subset \mathbb{R}^d$ are the time stamps and state values of a dynamically adapting stochastic process, respectively.

\subsection{Multi-task relatedness}
\label{sec:task_relatedness} 
This category induces biases to enhance multi-task learning in NPF members. 
Although the NPs naturally enable knowledge transfer by encoding multiple datasets as distributions over functions, such datasets are often sampled from the same i.i.d. distribution. The i.i.d. samples  reflect little on  real-world scenarios where, a learning agent (e.g., an autonomous driving engine) can be presented with multiple tasks (e.g., driving straight or with  turns, with or without dynamic obstacles, etc.) with the context of different tasks  sampled from distinct distributions (e.g., different weather conditions). The methods in this NPF branch thus target  multi-task learning in the face of tasks derived from such heterogeneous distributions.
Formally, these methods learn  $L$  training tasks such that the data for each task $l \in [1, L]$ is divided into a context set $C^l$ and a target set $T^l$. To reflect the heterogeneity in distributions, each task $l$ can be seen to be generated by a  function $f^l: \mathcal{X}_l  \rightarrow \mathcal{Y}_l$.

Requeima \textit{et al.} \cite{requeima2019fast} propose the  Conditional Neural Adaptive Process (CNAP) to balance the trade-off faced by few-shot multi-task classifiers while adapting to new tasks, \textit{i.e.}, striking a balance between model capacity and reliability  based on the number of parameters adapted to each task while still being resource efficient during the adaptation. CNAPs model CNPs as a combination of a task-specific adaptation model $\psi^l = \psi_\phi(.)$  with global parameters $\phi$ shared across tasks and a classification model with another set of pre-trained global parameters $\theta$.  Such a formulation  offers CNAPs a strong generalization capability to active learning and continual learning setups.

A more recent  formulation for inducing multi-task learning in NPs is the hierarchical Bayesian framework \cite{kim2022multitask, shen2022multitask, anonymous2023npcl, anonymous2023episodic}.
The resulting frameworks leverage a global latent that captures cross-task correlation and use it to condition the learning of per-task latents that capture finer task-specific stochastic factors. The global latent variable  thus serves as a natural enabler of knowledge transfer across  the low-level task-specific latents. 

\subsection{Domain Invariance}
\label{sec:domain_invariance}

This category targets learning more general representation of inputs with an immediate downstream application being the ability to distinguish the inputs belonging to similar domains.
It is further desirable that the learned representations remain task-agnostic to be employed to a wider range of downstream  applications. A popular choice for such learning objectives for NPs  are  the self-supervised contrastive losses \cite{chopra2005learning} that compare the similarity across the  representation of inputs. 
NPs can learn contrastively not only over  data points but also over the functional spaces. The latter objective, popularly known as functional contrastive learning (FCL) \cite{gondal2021function} thus helps bring the representations from the same function closer and that from other functions further apart.\footnote{While the first formal proof of FCL was put forward by Gondal \textit{et. al.} \cite{gondal2021function}, they do not consider using efficient regression techniques provided by NPs for richer representation learning. On a similar note, Mathieu \textit{et al.} \cite{mathieu2021contrastive} also exploit self-supervised  contrastive learning for NPs but they do away with exact reconstruction of posterior samples. We thus consider discussing these works from the perspective of future research directions (see Section \ref{sec:future_dir}).} 
Among the first applications of FCL includes Gao \textit{et al.} \cite{gao2022matters} using it to train CNPs for large scale vision-regression tasks.

Ye \textit{et al.} \cite{ye2022contrastive} further leverage contrastive learning to address the limitations of CNPs at jointly optimizing \textit{in-instantiation} observation prediction, \textit{i.e.,} generalization among the samples of a function and \textit{cross-instantiation} meta-representation adaptation, \textit{i.e.,} generalizing across a family of functions in the face of high-dimensional and noisy time series data. In particular, they highlight three major drawbacks arising from the CNP's optimization objective: (a) lack of predicitive correlations (see Section \ref{sec:pred_corr}), (b) limitations of generative models in forming abstractions for high-dimensional observations \cite{kipf2019contrastive}, and (c) supervision collapse in meta-learning \cite{doersch2020crosstransformers} due to entangled prediction and transfer tasks. To address these, they propose extending CNPs using an in-instantiation temporal contrastive learning (TCL) loss that aligns predictions with encoded ground-truth observations besides the cross-instantiation FCL loss that guides learning of meta-representation for each task.  

Contrastive losses are known to benefit from data augmentations \cite{chen2020simple}.
 While it is easy to employ geometric transformations for augmenting  vision-based signals \cite{oord2018representation}, time series domains such as audio processing, financial forecasting, etc. demand for tailored augmentation pipelines. To learn the representations for these domains using augmentations  that are generalizable across  domains, Kallidromitis \textit{et al.} \cite{kallidromitis2021contrastive} propose the Contrastive Neural Processes (ContrNPs) that exploit the fact that the aggregated encodings of an NP is shared across the target set while modeling a distribution over  functions. Using  forecasting in NPs as a supervised signal for unsupervised learning can therefore help generate  augmentations of a data point if the sampling functions are distinct.  ContrNP thus consists of a forecasting component based on NPs and a learning component based on the choice of self-supervised contrastive loss. 

  \section{Other domain-specific advances}
\label{sec:applications}
 While  section \ref{sec:classification} makes up the mainstream theoretical NPF advances, we identify a yet another track of research advances in the family that leverages the theoretical underpinnings  to deal with domain-specific problems. In what follows, we discuss five such major domains of NP applications with multiple existing works:  \textit{recommender systems}, \textit{hyperparameter optimization}, \textit{neuroscience}, \textit{space science}, and \textit{physics-informed modeling} of dynamic systems. We then brief  other such application areas which have seen relatively sparse applications of the NPF members. We categorize the latter as \textit{miscellaneous}.

\textbf{A. Recommender systems. } NPs have been employed to address the cold-start problem in recommender engines  by performing gradient updates for parameter initialization without considering the  recommendations of different users. For instance, the  Task adaptive Neural Process (TaNP) \cite{lin2021task} assumes each  task (user) to be an instantiation of a stochastic process and adapts to these based on a latent variable structure, a customized module and an adaptive decoder.  More recently, CNPs \cite{garnelo2018conditional} have been applied to collaborative filtering \cite{wang2022conditional}. Similar to TaNP, the conditional collaborative filtering process (CCFP)  predicts recommendations for partial targets,\textit{ i.e.,} the user-item interaction data. 

\textbf{B. Hyperparameter optimization. } Hyperparameter optimization (HPO)  involves selecting the optimal combination of hyperparameters for maximizing a model's performance. In real-world scenarios,  machine learning models often represent physical systems that are to be optimized. A naive way to decide their hyperparameters is to carry an exhaustive trial of the whole training process using the evaluated values (obtained from an acquisition function such as the Expected Improvement EI measure) of the chosen hyperparameters. However, given that each such trial can bear enormous costs (consider calibrating a power grid system each time a generator is added \cite{shangguan2021neural}), the widely-used sequential model-based optimization (SMBO) branch of HPO considers leveraging cheap-to-evaluate surrogate models. A Bayesian approach to SMBO further deals with querying the distribution (over functions) defined by the surrogate model, with the chosen hyperparameters. 

Given the success of GPs \cite{swersky2013multi} and neural networks \cite{snoek2015scalable} as choices for surrogate models in Bayesian SMBO, NPs stand as strong  candidates for  modeling  HPO.  Motivated by this, Luo \textit{et al.} \cite{9186363} propose the multitask CNP (MTCNP) where, GPs are replaced by CNPs as the choice of surrogate models for multi-task learning of a set of related optimization problems. The MTCNP model namely incorporates a correlation learning layer allowing the sharing of information among multiple task-specific CNPs. Similarly, Shangguan \textit{et al.} \cite{shangguan2021neural} propose the Neural Process for Bayesian Optimization (NPBO) framework to automate HPO of black-box power systems whose knowledge of internal workings is assumed to be unknown.  NPBO uses the EI measure as acquisition function and  replaces GPs with NPs for better scalability and accuracy in higher dimensional spaces. Wei \textit{et al.} \cite{wei2021meta} highlight the inaptness of the existing GP-based surrogate models for HPO in the context of SMBO based on transfer learning of hyperparameter configurations.  As a solution, they propose the Transfer Neural Process (TNP) to meta-learn the joint transfer of observations. 

\textbf{C. Neuroscience. } The meta-learning ability of NPs is useful to model the   responses of the visual cortex neurons to novel stimuli in a few-shot setting. Cotton \textit{et al.} \cite{cotton2020factorized} point out the limitation of NPs at learning a neuron's tuning function space in  $K-$shot regression  of stimulus-response samples. Inspired by the success of CNNs incorporating a factorized readout between the tuning function's location and properties on the  task \cite{klindt2017neural}, the authors  propose factorized NPs that host a factorized latent space obtained by stacking multiple NPs and passing the latent  variable computed by earlier layers to deeper layers. 
Pakman \textit{et al.} \cite{pakman20a} study NPs  for neural spike sorting, \textit{i.e.,} the grouping of spikes into clusters such that each cluster reflects the activity of different putative neurons. To do so, they propose the neural clustering processes (NCPs) that leverage spike waveforms as inputs, and are trained using $N$ context set points  and their $K < N$ distinct cluster labels. 

 \textbf{D. Space science. } Extraterrestrial data is usually captured  under challenging and uncertain environments (\textit{e.g.}, snaps of planets taken by a space probe during a brief flyby), and are typical of voids and irregularities. While interpolating  can offer a possible solution to filling such voids \cite{reuter2007evaluation}, interpolation relies on spatial continuity and thus produces values that often mirror the local information while ignoring the global topological patterns. On the other hand, a simple application of deep generative models like GANs \cite{goodfellow2014generative} might fill the void smoothly but do not provide any uncertainty estimates. NPs  carry the potential to offer the best of both worlds for such applications.

 Park \textit{et al.} \cite{park2021neural} employ  a sparse variant of ANPs \cite{kim2018attentive} for the probabilistic reconstruction of no-data gaps in the digital elevation maps (DEMs) of the moon's surface captured by narrow-angle cameras (NACs). The variant randomly samples $K$ points from the original attention window  which 
 are used to compute attention weights for a given context point $x_i$. The proposed sparse ANP thus reduces the computational complexity of ANPs from linear to a constant multiple of the squared dataset size. 
Similarly, {\v{C}}vorovi{\'c}-Hajdinjak \textit{et al.} \cite{vcvorovic2022conditional} consider CNPs \cite{garnelo2018conditional} for   modeling  stochasticity in the optically variable light curves  of active galactic nuclie (AGN) while 
Pondaven \textit{et al.} \cite{pondaven2022convolutional} employ ConvCNPs \cite{Gordon2020Convolutional} and ConvLNPs \cite{foong2020meta} to study the problem of inpainting missing pixel values in satellite images.  A peculiarity of the latter lies in using the Multi Scale Structural Similarity metric \cite{wang2003multiscale} as a choice of log-likelihood function for generating sharper mean predictions of images. 
 
 \textbf{E. Physics-informed systems design.} Inclusion of a priori physics-informed knowledge into the design of systems modeling real-world dynamic entities can help improve data efficiency, generalization as well as interpretability \cite{karniadakis2021physics}. For instance, Jacobian matrix in a robot manipulator provides the relation between joint velocities and end-effector velocities.  Inspired by this, Chakrabarty \textit{et al.} \cite{chakrabarty2021attentive} use ANPs to model climate-related impact of energy consumption in buildings by calibrating the digital twins of their grid-interactive simulations. These twins  incorporate the physics of heating, ventilation, and cooling (HVAC) and therefore, host interpretable outputs and parameters. 

 Other than HVAC physics, another widely used method for physics-informed modeling is  including the corresponding Ordinary Differential Equations (ODEs) into the learning problem's cost function.  While Norcliffe \textit{et al.} \cite{norcliffe2021neural}  apply neural ODEs \cite{chen2018neural} to model temporal dependency in data (see Section \ref{sec:temp_correlation}), ODEs further host a multitude of applications in modeling continuous-time real-world phenomena \cite{ahmed2021mathematical}. Prior to NPs, GPs have been successfully applied to  stochastic differential equation (SDE) modeling where they forecast the behavior of  continuous-time systems \cite{garcia2017nonparametric}. Motivated by this, Wang \textit{et al.} \cite{wang2021neural} consider NPs to model  stochastic differential equations (SDEs)  defining forward passes in DNNs as state transformations of a dynamical system. The resulting problem setup is thus capable of estimating epistemic uncertainty in DNNs using real valued continuous-time stochastic processes like the Wiener process \cite{oksendal2003stochastic}. More recently, Wang \textit{et al.} \cite{wang2022np} proposed using NPs as stochastic surrogates for modeling ODEs defining the simulation of complex nonlinear systems through finite element analysis (FEA) \cite{zienkiewicz2005finite}. 
 
 \textbf{F. Robotics. } With the ubiquitous spread of automation in machines, uncertainty-aware predictions in these are crucial for their real-world deployment.  NPs are thus handy to the domain of uncertainty-aware robotic task modeling. Chen \textit{et al.} \cite{chen2022meta} employ CNPs for robotic grasp detection of unseen objects. 
  Li \textit{et al.} \cite{li2022category} leverage CNPs on the upstream task of 6D pose estimation that can enable a robotic arm to be aware of the position and orientation of objects in its vicinity. 
Similarly, Yildirim \textit{et al.} \cite{yildirim2022learning} introduce CNPs to robotics in the context of learning social aspects of navigation. Here, CNPs  first generate navigation trajectories of a given environment followed by the generation of goal-directed behaviors that take into account the environment's social elements such as the presence of pedestrians on a footpath. 
 
 \textbf{G. Miscellaneous.} While the aforementioned domains comprised works from multiple research directions, we find some further budding NPF application domains that  contain work done in isolation. A summary of few such works together with their target domains is detailed in Table \ref{table:applications}. As the NPF members enjoy increasing popularity across a range of real-world application domains, we expect the entries in this table to grow  with a number of such related entries eventually forming their own application domain. 
 
 \begin{table*}[]
 \centering
 \footnotesize
  \caption{A summary of budding application domains of the NPF members.}
\begin{tabular}{p{5.7cm}@{\hskip 0.1in}p{10cm}@{}}
\toprule
\multicolumn{1}{c}{Domain [Author]}  & \multicolumn{1}{c}{Summary}  \\ 
\midrule

 Detecting psychiatric disorder biomarkers in clinical fMRI scans \cite{kia19a}  & Uses latent NPs for the normative modeling \cite{marquand2016understanding} of the spatially structured mixed-effects in functional magnetic resonance imaging (fMRI) formulated using tensor Gaussian predictive process \cite{kia2018scalable}.\\ \cmidrule{1-2}
  Interactive Attention Learning (IAL) for improving interpretability of neural networks \cite{heo2020cost} & Proposes the IAL framework for generating cost-effective attention guided by
an active human-in-the-loop correction of a neural network’s attention interpretations. Exploits latent variable NPs to model the context
set whose human-generated attention annotations are given. \\ \cmidrule{1-2}
  Editable training of neural networks to correct prediction mistakes \cite{Sinitsin2020Editable} & Proposes adding condition vectors to intermediate
activations of CNPs to reduce the effect of catastrophic forgetting in editable training by minimizing the KL divergence between the predictions of original and edited models. \\ \cmidrule{1-2}
   Statistical downscaling of climate model outputs \cite{vaughan2021convolutional} &  Exploits ConvCNPs to address the limitations of existing statistical multi-site downscaling methods, namely training on off-the-grid climate data and predicting on  locations unseen during training. \\ \cmidrule(lr){1-2}
 Unsupervised detection of anomalous machine sounds \cite{wichern2021anomalous} & Uses ANPs to define priors over log mel spectograms of normal sounds thus bypassing the need for pre-specified masked regions \cite{suefusa2020anomalous} over these. \\ \cmidrule{1-2}
 Fake news detection for emergent events with few verified posts \cite{wang2021multimodal} & Employs ANPs with a hard-attention mechanism \cite{jang2016categorical} to select relevant context out of class imbalanced fake news context datasets.  \\ \cmidrule{1-2}
Predictive autoscaling of computing resources in the cloud \cite{xue2022meta}   &  Employs ANPs to meta learn the unit workload to CPU utilization mapping in a model-based reinforcement learning setting that seeks the optimal numbers of virtual machines in the cloud to maintain the CPU utilization is at a target level.\\ \cmidrule{1-2}
 Multi-fidelity surrogate modeling of epidemiology and climate modeling tasks \cite{Wu2022MultifidelityHN}  & Leverages latent NPs to build Multi-fidelity Hierarchical Neural Processes (MF-HNP) that learn the joint distribution of high and low-fidelity outputs. MF-HNP fuses data with varying input and output dimensions at different fidelity levels. Training involves inferring two latents, one for each fidelity level.  \\ \cmidrule{1-2}

  Semi-supervised image classification  of unlabeled points conditioned on labeled ones \cite{wang2022npmatch} & Uses an NP for incrementally obtaining noise-free predictions on unlabeled target points; introduces a realistic test setup for NPs by storing training features in memory banks and retrieving them as test time context.\\ \cmidrule{1-2} 

 Compositional law parsing with concept representation for scenes \cite{shi2022compositional} & Applies latent NPs for encoding concept-specific latent variables from each context image. Following the  standard NP decoder setup, these variables are concatenated with the queried target image to predict the target concepts.\\
\bottomrule
 \end{tabular}

 \label{table:applications}
\end{table*}

\section{Datasets}
\label{sec:datasets}
Table \ref{table:datasets} shows the diversity of the fields that the NPF members have been successfully applied to along with their associated datasets and  properties. 
The key to modeling the stochastic processes generating these datasets lies in the appropriate problem formulation. For instance, the MNIST digits \cite{lecun1998gradient} can be employed in the context of regression as well as classification based on whether we choose to model the stochastic process generating the distribution of pixel intensities or the distribution of digit labels \cite{garnelo2018conditional}. Similarly, the Omniglot dataset \cite{lake2015human} with fairly large number of classes and few instances per class suits well for one-shot classification and for image in-painting. 
\begin{table*}[]
\centering
\scriptsize
\caption{Table showing the diversity of applications, their associated datasets, and their first usage in the context of NPF members.}
\begin{tabular}{p{3.5cm}@{}p{4cm}@{}p{2.5cm}@{}@{}p{1cm}@{}p{2cm}@{}p{2cm}}
\toprule
 \multicolumn{1}{c}{Domain}  & \multicolumn{1}{c}{Dataset}  &  \multicolumn{1}{c}{First usage} & \multicolumn{1}{c}{Dimensions}  & \multicolumn{1}{c}{Size} &  \multicolumn{1}{c}{Continuum}   \\ \midrule
 \multirow{5}{3.4cm}{\centering Image completion} 
 &MNIST  \cite{lecun1998gradient} & \multirow{2}{2.4cm}{ \centering \cite{garnelo2018conditional}} & \text{2-d} & & Spatial  \\ 
 & CelebA \cite{liu2015faceattributes} &  & 2-d & & Spatial\\ \cmidrule{2-3}
 & EMNIST \cite{cohen2017emnist} & \cite{lee2020bootstrapping} & 2-d & & Spatial \\ \cmidrule{2-3}
 & SVHN \cite{netzer2011reading} & \multirow{1}{2.4cm}{ \centering \cite{Gordon2020Convolutional}} & 2-d & & Spatial \\
\midrule
 
 \multirow{2}{3.4cm}{\centering Higher dimensional image completion} & Rotating MNIST \cite{casale2018gaussian} & \cite{norcliffe2021neural} & 3-d & & Spatio-temporal \\ 
  & Synthetic bouncing ball \cite{sutskever2008recurrent} & \cite{ye2022contrastive} & 3-d & & Spatio-temporal \\ \midrule
 \multirow{1}{3.4cm}{\centering One-shot classification} & Omniglot \cite{lake2015human}& \cite{garnelo2018conditional} & 2-d & 32,460 & Spatial \\ \midrule
 \multirow{2}{3.4cm}{\centering Bandit formulation of paraphrase identification} & Quora Question Pairs\footnote{\url{https://quoradata.quora.com/First-Quora-Dataset-Release-Question-Pairs}} & \multirow{2}{2.4cm}{\centering \cite{weber2018optimizing}} & & 400,000 & Text \\
 & MSR corpus \cite{dolan2004unsupervised} & & & 5,800 & Text \\ \midrule
 \multirow{1}{3.4cm}{\centering Predator-Prey  modeling} & Hudson’s Bay hare-lynx data \cite{goel1971volterra} & \cite{Gordon2020Convolutional} & & & \\ \midrule
 \multirow{1}{3.4cm}{\centering Photometric time series} & PLAsTiCC \cite{allam2018photometric} & \cite{Gordon2020Convolutional} & 2-d & 3,500,734 & Temporal \\ \midrule
 \multirow{3}{3.4cm}{\centering User cold-start recommendation} & MovieLens-1M\footnote{\url{https://grouplens.org/datasets/movielens/1m/}} & \multirow{3}{2.4cm}{\centering \cite{lin2021task}} & & & \\
 & Last.FM\footnote{\url{https://grouplens.org/datasets/hetrec-2011/}} & & & & Text \\
 & Gowalla \cite{cho2011friendship} & & & & \\ \midrule
 \multirow{2}{3.4cm}{\centering Traffic speed modeling} & PEMS-BAY \cite{wu2019graph} & \multirow{2}{2.4cm}{\centering \cite{yoo2021conditional}} & &  16,937,179 &  \\
 & METR-LA \cite{jagadish2014big} & & &  7,245,000 & \\ \midrule
 \multirow{3}{3.4cm}{\centering Noisy time series modeling} & AFDB \cite{moody1983new} & \multirow{3}{2.4cm}{\centering \cite{kallidromitis2021contrastive}} & 2-d & & Temporal \\
 & IMS Bearing \cite{qiu2006wavelet} &  & 2-d & & Temporal \\
 & Urban8K \cite{salamon2014dataset} &  & 2-d & 8,732 & Temporal \\ \midrule
 \multirow{1}{3.4cm}{\centering Precipitation modeling} & ERA5-Land \cite{munoz2021era5} & \cite{foong2020meta} & 3-d & & Spatio-temporal \\ \midrule
 \multirow{1}{3.7cm}{\centering Physics-informed modeling} & Cart-Pole simulator \cite{gal2016improving} & \cite{wang2020doubly} & & Variable & \\ \midrule
 \multirow{1}{3.4cm}{\centering Graph data modeling} & Graph benchmark datasets \cite{KKMMN2016} & \cite{carr2019graph} & & Variable &  Graphs \\ \midrule
 \multirow{3}{3.4cm}{\centering Graph link prediction } & Cora \cite{mccallum2000automating} &  \multirow{3}{2.4cm}{\centering \cite{liang2021neural}} & & 2,708 nodes & Graphs \\ 
  & Citeseer \cite{giles1998citeseer} & & & 3,327 nodes & Graphs \\ 
  & PubMed \cite{sen2008collective} & & & 19,717 nodes & Graphs \\ \midrule
  Point cloud part labeling & ShapeNet \cite{chang2015shapenet} &  \cite{DBLP:journals/corr/abs-2009-13895} & 3-d & & Sptaial \\ \midrule
  \multirow{1}{3.4cm}{\centering Continual Learning} & Split-MNIST \cite{zenke2017continual}, Split-CIFAR100 \cite{chaudhry2018riemannian} & \cite{requeima2019fast} & 2-d & & Spatial \\ \midrule
  \multirow{1}{3.4cm}{\centering Active Learning} & Flowers \cite{Nilsback08}, Omniglot \cite{lake2015human} & \cite{requeima2019fast} & 2-d & & Spatial \\
\midrule
\end{tabular}
\label{table:datasets}
\end{table*}

\section{Depiction of Function Modeling}
\label{sec:task_nature}
This section shows the range of tasks that can be targeted using the NPF members. In particular, we categorize the tasks where, the domain of the function $f: \mathcal{X} \rightarrow \mathcal{Y}$  can be situated in one ($\mathcal{X} \in \mathbb{R}$), two ($\mathcal{X} \in \mathbb{R}^2$), or three ($\mathcal{X} \in \mathbb{R}^3$) dimensional spaces. We visualize and discuss a representative task of each such category. On 1d and 2d inputs, we leverage the previously established regression frameworks for our experiments. On 3d inputs, we adapt a graph-based NP to perform ShapeNet part labeling. It is worth noting that the motivation behind our experiments remains showing the nature of tasks that NPs can model rather than benchmarking their performances. We discuss further implementation details and results in App. \ref{app:task_nature}.

\subsection{1-d function modeling}
\begin{figure}[t!]
\centering
\includegraphics[width=0.5\textwidth]{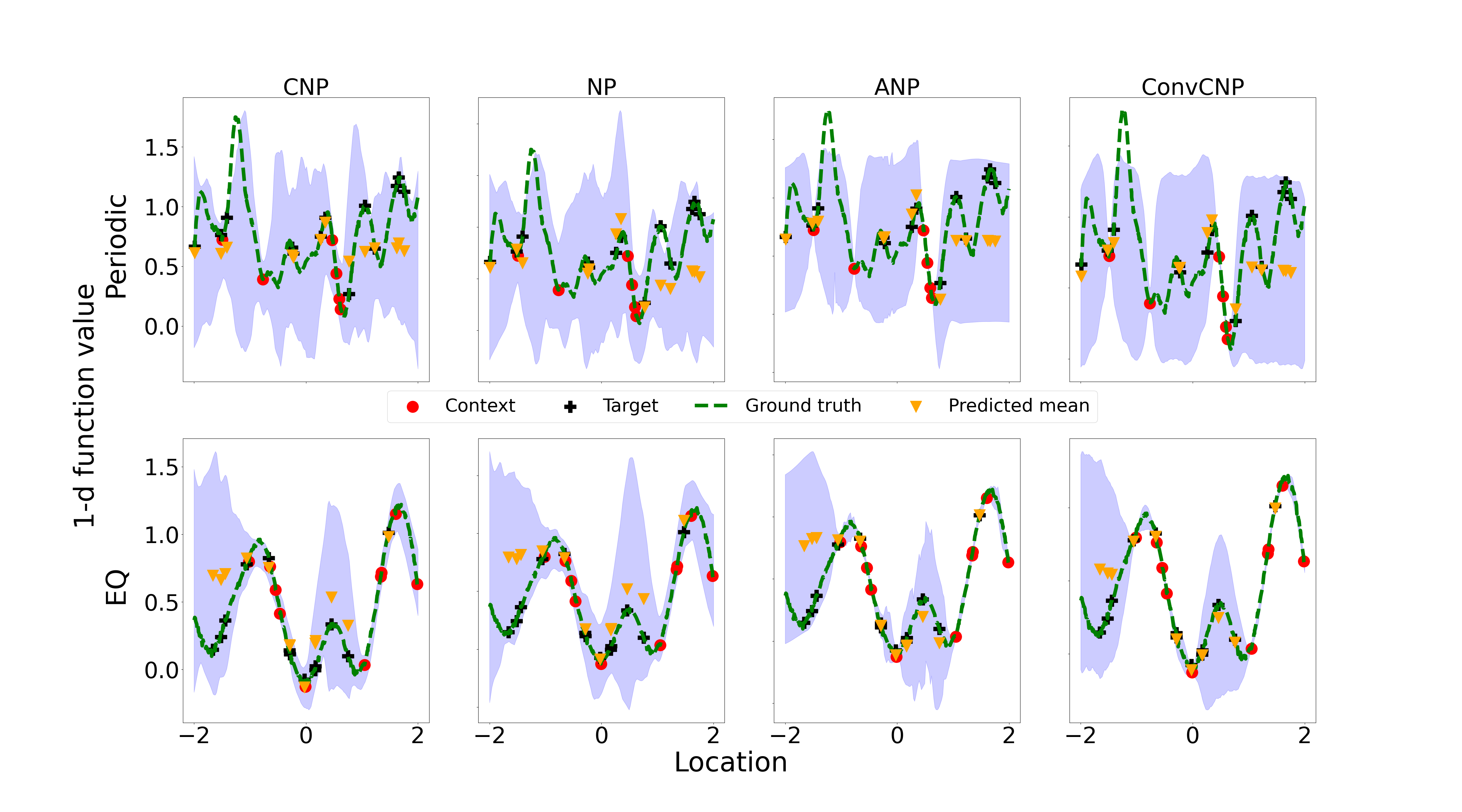}
\caption{The mean and variance of the CNP, the NP, the ANP, and the ConvCNP on the 1-d regression task of predicting the values at the target locations of a ground truth Gaussian process (dashed green lines) based on periodic (top row) and EQ (bottom row) kernels.}
\label{fig:gp_exp}
\end{figure}

For 1-d function modeling, we first generate two datasets using GPs with a static exponential quadratic (EQ) and a periodic kernel. We then train NPs to regress the value of these kernel functions at the locations provided.  For evaluation, we interpolate the models by randomly choosing the number of context points in the range $[5,20]$ while fixing the number of target points at 13. Fig. \ref{fig:gp_exp} shows the mean and variances of   four common NP variants: the CNP \cite{garnelo2018conditional}, the NP \cite{garnelo2018neural}, the ANP \cite{kim2018attentive}, and the ConvCNP \cite{Gordon2020Convolutional} on the target set of data generated by the  kernels. While all the models fit the periodic data to a good extent, on the EQ kernel, the robust inductive biases of ConvCNP and ANP offers them   richer predictive correlation modeling.

\subsection{2-d function modeling}
\label{sec:image_completion}
For 2-d inputs, we consider modeling the  RGB  image pixel generation task  on   the CelebA dataset \cite{liu2015faceattributes}. We use the same four NP variants as for 1-d regression  except for ConvCNPs which we replace with its latent  counterpart GBCoNP \cite{wang2021global} for crisper visualizations. We use the pretrained weights of Dubois \textit{et al.} \cite{dubois2020npf} for the CNP, the NP, and the ANP, and that of Wang \textit{et al.} \cite{wang2021global}  for the GBCoNP.

\begin{figure}[h!]
    \centering
      \includegraphics[width=0.45\textwidth]{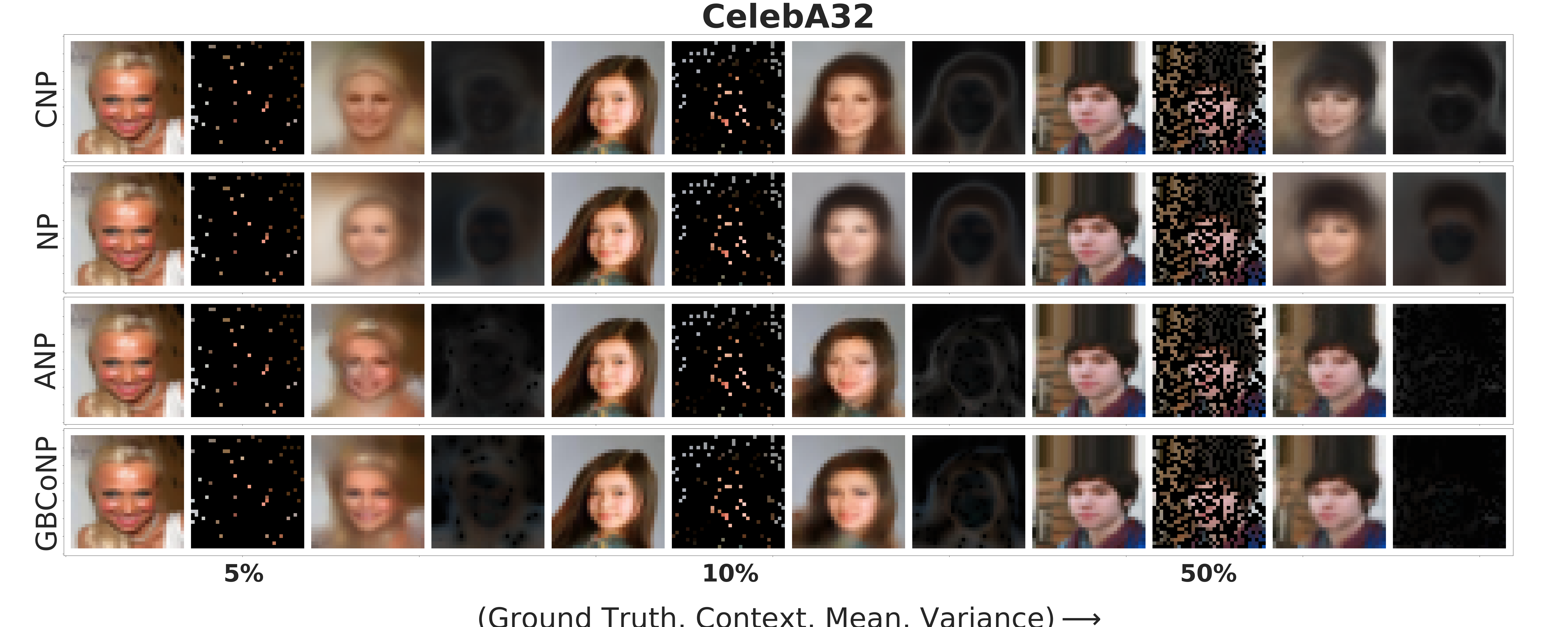}\\
\caption{Predictive mean, and variances of CelebA $32 \times 32$ images with  context sizes amounting to  5, 10, and 50\% of the total image pixels.}
\label{fig:imagecompletion}
\end{figure}
\begin{figure}[t!]
    \centering
        \includegraphics[width=0.3\textwidth]{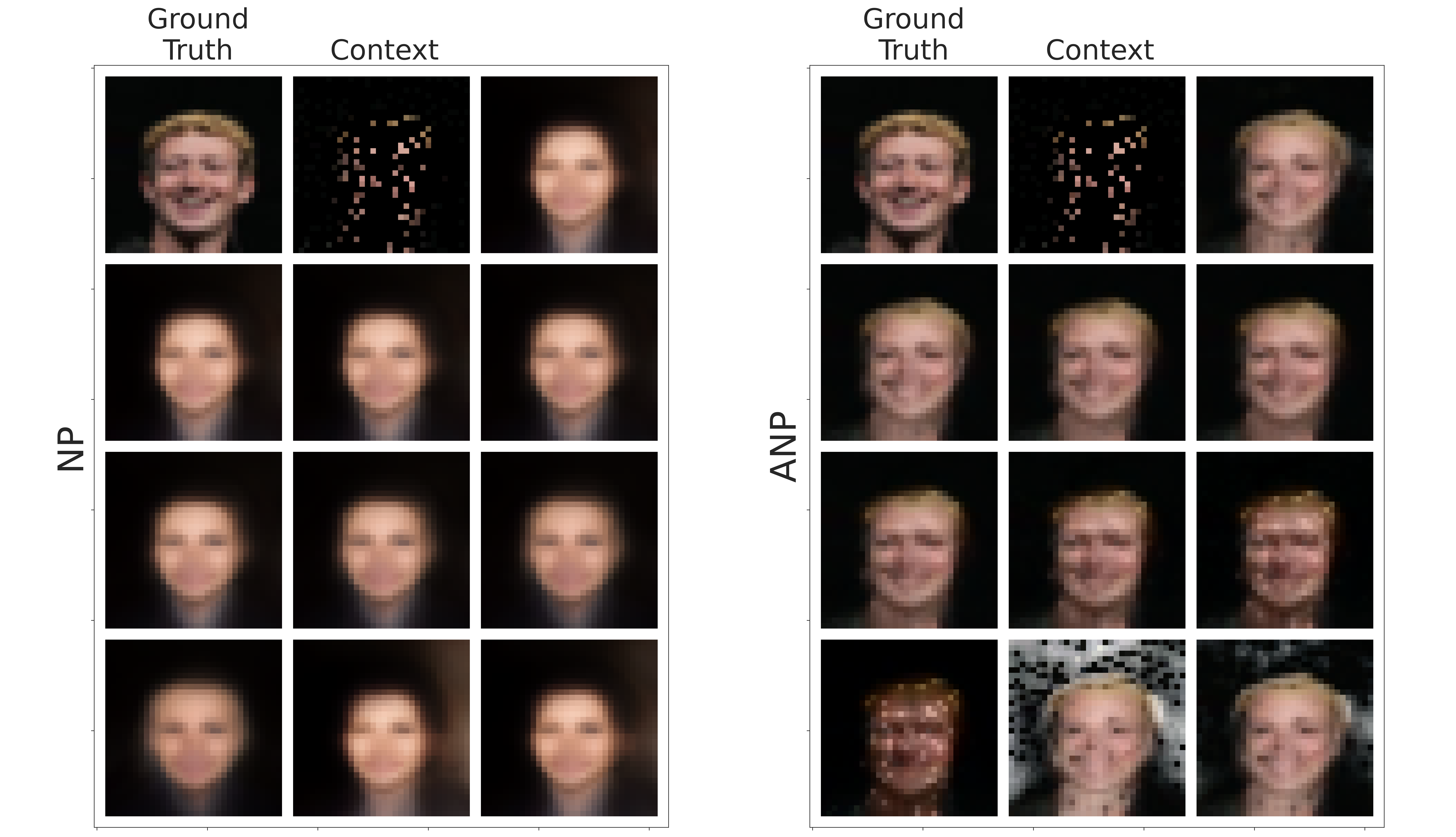}
    \caption{Depiction of varying predictions of the NP and the ANP over 10 distinct latent samples on CelebA-32. For reference, the first two  images in each grid represent the ground truth and the context, respectively.}
    \label{fig:globaluncertainty}
\end{figure}

 
Fig. \ref{fig:imagecompletion} visualizes the predictive mean and variances of each model by randomly selecting $5\%, 10\%,$ and $50\%$ of the total image pixels as the  context points and providing the locations of the entire image pixels  as targets. The mean predictions of each model can be seen to  grow crisper with more context.  Given that Celeb-A hosts  a majority of female faces, the CNP and NP have their mean predictions resemble closely to the average of all the faces, \textit{i.e.,} more feminine attributes.  To emphasize the role of prior in inducing coherence among the predictions, we visualize the outputs of the NP and the ANP corresponding to ten different  samples by fixing the context.  Fig. \ref{fig:globaluncertainty} shows  both the models generating a range of  predictions that are equally justifiable to the context. 

\subsection{3-d function modeling}
To model functions on 3d domains, we consider the classification task  of part labeling on the ShapeNet part dataset \cite{yi2016scalable} that hosts 50  part labels of 16 different objects.   Our NP architecture uses the dynamic graph CNN (DGCNN) \cite{wang2019dynamic}  as  the encoder and decoder modules  to induce  edge convolutions.  Inputs to the NP include a tuple of coordinate locations  $(x, y, z)$ and a 1d categorical descriptor of the object for the cloud's global representation. The goal is to predict the target set labels alongside their uncertainties (see App. \ref{app:3d_modeling} for implementation details).

\begin{figure}[t!]
    \centering
    \includegraphics[width=0.5\textwidth]{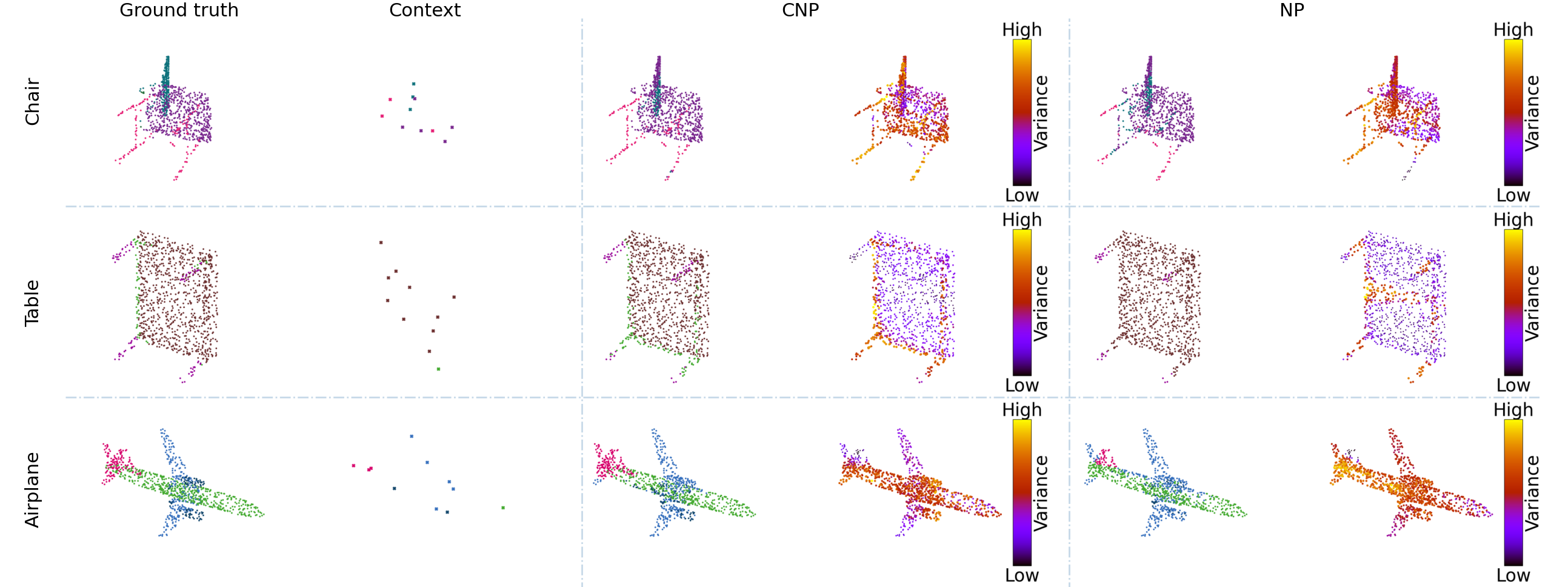}
    \caption{Visualization of part labeling. Each shape depicts 1024 sampled points which form (from left to right): ground truth part labels, 1\% context, and the predictive means and variances of the DGCNN-based CNP and NP models. For better visibility: (a) the size and color of the points  denote uncertainties, (b)  the context  points have been magnified  and  might differ in  alignment with the other shapes.} 
    \label{fig:pointcloud}
\end{figure}

\begin{figure}[t!]
\begin{minipage}{\textwidth}
\begin{tikzpicture}
  \node (img)  {\includegraphics[width=0.4\textwidth]{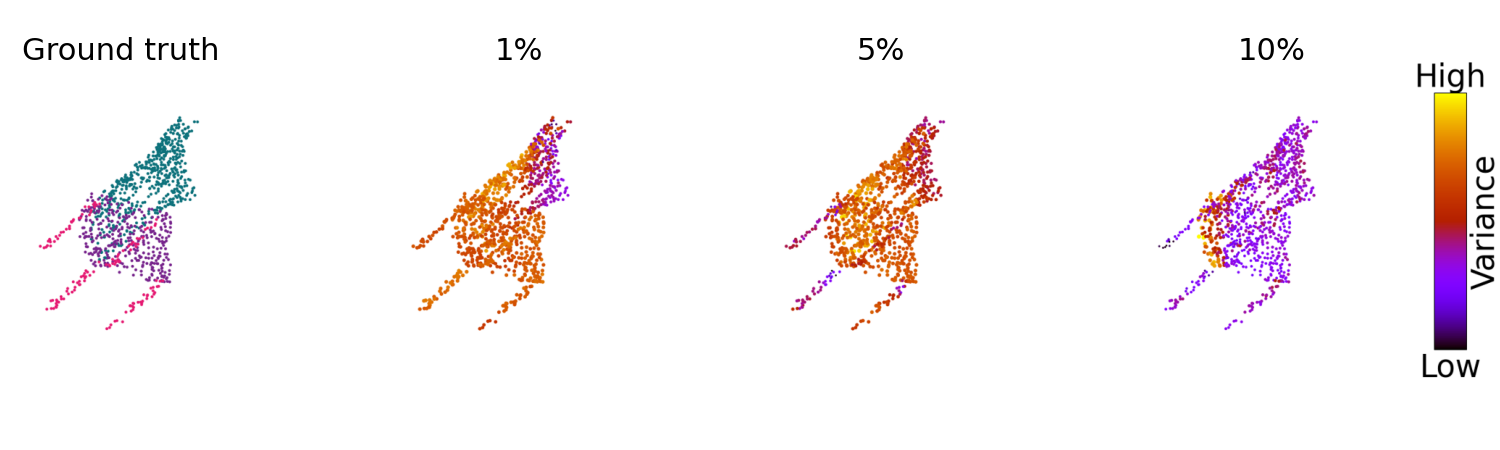}};
  \node[below=of img, node distance=0cm, yshift=1.5cm,font=\color{black}] {Context size $\rightarrow$};
\end{tikzpicture}
\end{minipage}%
\caption{Visualization of the predictive uncertainties of the CNP for  part labeling of the point cloud of a chair with varying context sizes. The sizes and colors of the points correspond to uncertainties: larger sizes towards yellow end of the spectrum denote a higher predictive variance.} 
\label{fig:varycontexts}
\end{figure}

 Fig. \ref{fig:pointcloud} shows
the mean and variance of the DGCNN-based CNP and NP models based on $1\%$ context, \textit{i.e.,} 10 observed points across 3 different categories:  chair, table, and airplane. Both models output more uncertain predictions for points lying at the junctions and boundaries of the shapes. The CNP without any probabilistic sampling can be seen to produce more accurate overall predictions than the NP.   The superior predictive results of the CNP stands in line with the previous works \cite{garnelo2018conditional, garnelo2018neural}.  To study the role of the  context, we ablate the performance of the CNP model by varying the context sizes.  Fig. \ref{fig:varycontexts} shows that as the size of the context  increases from $1\%$ to $10\%$ of the total points, the model gets more certain about its predictions of the rails, seat, stiles, and legs of the chair.

\section{Future Research Directions}
\label{sec:future_dir}
Uncertainty-aware modeling of stochastic processes has a huge potential for improvement despite the success of the NPF members over static kernel-based GPs. In this section, we outline some  common issues faced by the current NPF models and discuss the directions for future research. 
\par
 \textbf{Cost-efficient generalization. }    The identification of underfitting in NPs \cite{kim2018attentive} led to a number of follow-up works adding up the computational costs by considering robust Janossy pooling mechanisms such as self-attention. Such mechanisms however grow quadratically with the number of input points. One efficient solution to attention-based generalizability could be applying restricted attention mechanisms \cite{beltagy2020longformer} or inducing points methods \cite{wu2021hierarchical} that scale linearly with sequence length. Using locality-sensitive hashing \cite{zandieh2020scaling} and KD-trees \cite{shen2005fast} can further help bringing the cost down to the order of logarithmic scales of the sequence length. Moreover, as a low cost alternative to avoiding underfitting, it can be plausible to leverage the recent findings in data-efficient  meta-learning to the NP family. For instance, we expect to see the amount of supervision required by NPs to attain a desired level of generalization in the samples of the learned predictive distribution \cite{sun2021towards}.

Another promising direction for scalable generalization in NPs can be enhancing the latent variable representation using more flexible equivariant transformations \cite{holderrieth2021equivariant, kawano2021group} over the context for instance. Such group equivariances can in turn be efficiently scaled to higher dimensions by opting for techniques such as group equivariant subsampling operations \cite{xu2021group}. On a similar note,  scalablilty and accuracy can be jointly achieved by making the traditional gaussian processes fast enough to be incorporated in NPs for more accurate predictions on higher dimensional data \cite{shi2019scalable}.
\par

\textbf{Improved set representations. } Current  NP variants rely either on 1 or 2-order permutation instantiations of  Janossy pooling  (see Section \ref{sec:resemblances}). As such, efficient inducing of higher degree input relational reasoning  into the models remains an active area of research. For example, NPs could be applied to encode all $N!$ permutations of the context set and average only over randomly sampled subset of encodings to get a more fair estimate of the exact permutation invariant representation \cite{murphy2018janossy}. Alternately, an adversary could be used to find out permutations with maximum loss values such that the resulting  training objective  amounts to computing permutation-invariant representation that minimizes the adversarial impact \cite{pabbaraju2019learning}. An advanced step along this direction will be to consider applications modeling the distribution of sets where, each element is associated with its own symmetry, for example, sets of 3-d point clouds, graphs, etc. \cite{maron2020learning}.  Similarly, alternate choices for aggregation functions could be explored given that their sensitivity to the performance of Deep Sets \cite{soelch2019deep}.

\textbf{Better function priors. } Given that NPs posit a prior directly over the functional space occupied by the neural networks and that such priors might not always hold for deeper architectures, improving the function priors can serve as a key enabler for a full Bayesian treatment of Neural Processes. To this end, potential directions remain imposing manifold structure on the latent distribution \cite{falorsi2019reparameterizing}, using the interpretable priors of GPs as a reference to match with \cite{tran2021sensible}, and meta-learning of the priors \cite{harrison2018meta}. 

\par 
\textbf{Likelihood-free density estimation.} Modeling the  predictive distribution  requires NPs to rely on an explicit likelihood assumption for $p$. While the  assumption works fine for low dimensional inputs, modeling complex higher dimensional predictive distributions using explicit likelihoods require learning magnitudes of greater parameters. This in turn leads to   posterior beliefs being biased \cite{havasi2018inference}. Avoiding  biased posterior belief  requires the sampling procedure to be sequential thus adding to  computational costs. Using the sampling of high-dimensional images as an example, avoiding biased posterior means switching to the computationally challenging task of conditional image generation \cite{cope}. One way to efficiently recover the unbiased posterior belief for CNPs could be doing away with pixel-perfect reconstruction and instead learning context in a self-supervised manner \cite{mathieu2021contrastive}. However, there still remains a large room for improving the cost/data efficiency of such techniques. 
\par 
\textbf{Context sensitivity.} For tasks involving long-tail distributions of the ground truth data generating processes (such as imbalanced classification), the selection of context points during training can greatly affect the performance of the NPs. While a better measure to further evaluate the robustness of NPs can be to train these on different context and target set sizes  \cite{le2018empirical}, this could be computationally demanding. The NP research community should thus prioritize evaluating the robustness of their models to the quality of the training/testing context points besides their sizes. A  motivation for this can be drawn from other order-sensitive deep learning domains such as exemplar-based continual learning \cite{jha2021continual} wherein the criteria for selecting exemplars largely influences a model's performance \cite{rebuffi2017icarl}. 

\section{Conclusion}
In this paper, we surveyed the recent advances in the Neural Process Family (NPF). We first laid the foundation for  NPs and related it to a number of contemporary advances in other deep learning domains. We introduced a classification scheme that encompasses the up-to-date fundamental research in the domain. We then followed up by summarizing a number of key domains that NPF methods have been actively applied to. To the end goal of depicting the tasks which NPs can handle, we  visualized  the function modeling on 1-d, 2-d and 3-d inputs. Lastly, we briefed a few promising  directions that can address the existing limitations of NPF members. We hope this survey serves as an introductory material for readers from the academia and the industry, and helps accelerate the pace of research in the field.

\bibliographystyle{IEEEtran}
\bibliography{IEEEabrv,tpami}
 \clearpage

\appendices
\begin{figure*}[t!]
\centering
\subfloat[Conditional Neural Process (CNP)]{\includegraphics[width=2in]{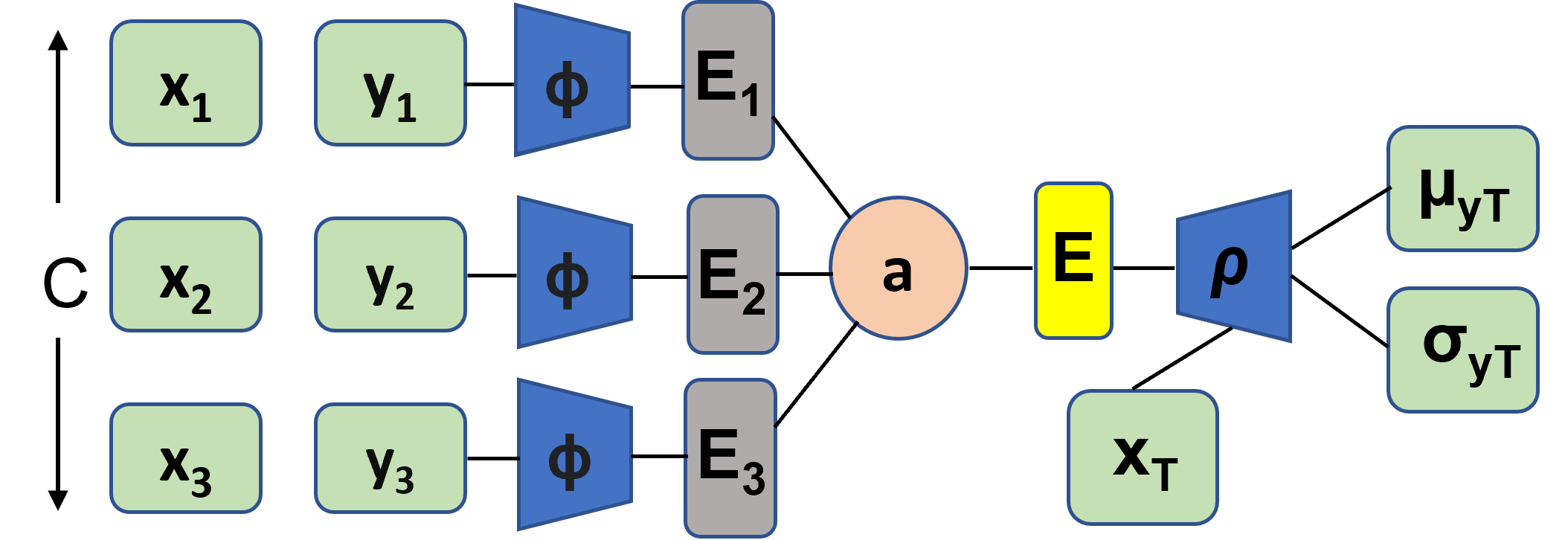}%
}
\hspace{0.5in}
\subfloat[(Latent) Neural Process (NP)]{\includegraphics[width=2.2in]{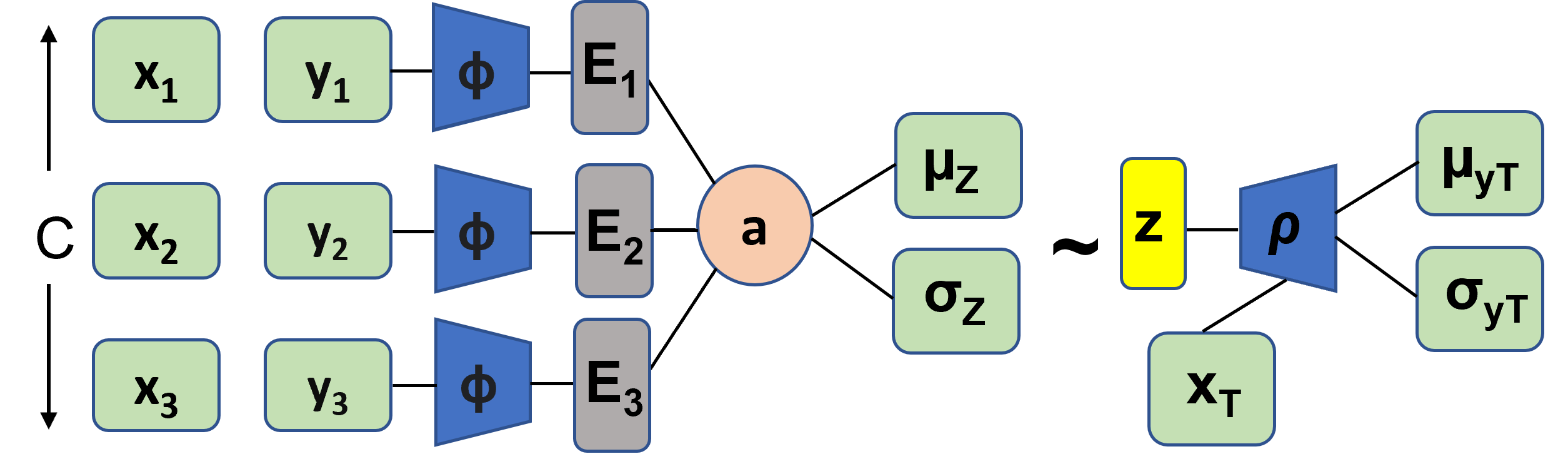}%
}

\caption{Illustration of Neural Process architectures adapted from \cite{garnelo2018conditional}: $C$ is the context set composed of three labeled data points $(x_i, y_i)$. $\phi$ is the encoder network acting on individual data points to produce encodings  $E_i$ which are aggregated by the operator $a$. $\rho$ is the decoder network conditioned on the target location $x_T$ and the aggregated context encoding. While the CNP feeds the output of $a$ directly to $\rho$, the NP first maps it to a distribution from which the latent variable $z$ is sampled to be fed to the decoder. $\mu$ and $\sigma$ denote the means and variances of the respective distributions.}
\label{fig:background}
\end{figure*}
\section{Stochastic processes}
\label{app:stochastic_process}
 Non-parametric stochastic processes have served as traditional alternatives to  presuming a prior distribution over the parameters of a function such as the weight priors in Bayesian neural networks (BNNs). A non-parametric process can be thought of capturing the data generating behavior and hence, presumes a prior over a family of functions that together summarize this behavior. However, unlike the case of BNNs with priors over a finite-dimensional Banach space (of parameters), assigning a distribution over infinite-dimensional Banach space (of functions) is not straightforward due to the lack of a Lebesgue measure \cite{lebesgue1902integrale} on the latter. To achieve the latter goal,  the Kolmogorov extension theorem \cite{oksendal2003stochastic} provides two necessary conditions for any collection of bounded distributions to mimic a stochastic process over functions. Assuming our dataset to be defined by a target set $T$ containing $M$ features $\{x_1, ..., x_M\}$ and their values $\{y_1, ..., y_M\}$ such that $T$ follows a well-defined continuous probability distribution. Then, for the joint (posterior) probability distribution of the values of a context set  $C \subset T$ comprising $N < M$ points to define a stochastic process modeled by a smooth, continuous bounded function $f$, the following two conditions must hold:
\begin{enumerate}
    \item Exchangeability:  it must remain unaffected by the output of the ordering function $\pi$ of its elements, \textit{i.e.,} $p_{x_{1:N}}(y_{1:N}) = p_{\pi(x_{1:N})}(\pi(y_{1:N}))$,
    \item Consistency: it must remain unaffected upon marginalizing arbitrary number of entries of the predictions, \textit{i.e.,} $p_{x_{1:N}}(y_{1:N}) = \int p_{x_{1:M}}(y_{1:M}) dy_{N+1:M}$ where, $1 \leq N \leq M$.
\end{enumerate}

\subsection{Gaussian processes.} 
\label{app:gp}
One widely used instance of stochastic processes that adhere to the aforementioned criteria are the Gaussian processes (GPs) \cite{williams2006gaussian}. GPs  restrict $T$ to follow a multivariate Gaussian distribution through the use of a  kernel function $k$ that captures the covariance structure among the locations. In the limited data regime, GPs thus offer two major benefits over BNNs as  a framework for predictive uncertainty measurement: 1) specifying the prior knowledge analytically in the form of a kernel function that exploits the covariance structure among the datapoints helps them encode any inductive bias into the model, and 2) inference in GPs for a query point $x^* \in T$ involves computing the posterior mean $\mathbb{E}(f(x^*)|f(C))$ as a linear combination of the kernel function values $k(x^*, x); x \in C$ and is thus much simpler than dealing with the intractability of priors as in BNNs. Moreover, unlike BNNs, the GP approach is non-parametric (more intuitively, \textit{infinitely parameterized}) in that it finds a distribution over the possible infinite dimensional function spaces that are consistent with the observed data.  
\par
\textbf{Limitations of GPs.}  GPs use the entire sample information in computing the posterior mean and thus lack sparsity. Such dense information leaves standard GPs with a number of setbacks \cite{binois2021survey}: (a) costly training and inference procedures due to the posterior mean function scaling cubically in the number of observations $N$, (b) difficulty in covariance computation when exposed to high dimensional feature spaces due to the distance between uniformly sampled points concentrating
increasingly further away, and (c) data inefficiency due to the challenging notion of encoding prior knowledge for kernel specification. Moreover, designing appropriate analytical priors can be a hard task on its own. While a range of works address these issues by trading one or more metric(s) (for example, correlation \cite{quinonero2005unifying}) for computation, constructing scalable GP kernels still remains a domain of active research. On the other hand, deep neural networks ensure such scalability of predictive quality with the number of data points.  
The latter thus make a good alternative to overcome the challenges of GPs.

\section{Function Modeling Experiments}
\label{app:task_nature}
In this section, we discuss the range of tasks that can be targeted using the NPF members. In particular, we categorize the tasks where, the function $f: \mathcal{X} \times \mathcal{Y}$ modeling the  stochastic process can be a function of one ($\mathcal{X} \in \mathbb{R}$), two ($\mathcal{X} \in \mathbb{R}^2$), or three ($\mathcal{X} \in \mathbb{R}^3$) dimensional inputs. For empirical affirmation, we conduct experiments on a representative task of each such category using NP baselines and discuss the implications of the results. On 1d and 2d inputs, we leverage the previously established regression frameworks for our experiments. On 3d inputs, our work marks the first step for ShapeNet part labeling using NPs. It is worth noting the motivation behind our experiments remains pointing out the nature of tasks that NPs can address rather than benchmarking their performances.

\subsection{1-d function modeling}
For 1-d function modeling task, we first generate two datasets using GPs with a static exponential quadratic (EQ) and a periodic kernel. We then employ NPs to regress the value of these kernel functions at the locations provided. To do so, we train our models by sampling a curve from the GP with a variable number of context and target points. Each model is trained for $2e^5$ iterations. For evaluation, we interpolate the models by randomly choosing the number of context points in the range $[5,20]$ while fixing the number of target points at 13. Following the standard evaluation protocols, we use the negative log likelihood (NLL) loss to report the predictive log-likelihood and mean squared error (MSE) to report the reconstruction error of the models.

Table \ref{table:1dscores} shows the mean and standard deviation of the NLL and MSE of the four basic NP variants: the CNP \cite{garnelo2018conditional}, the NP \cite{garnelo2018neural}, the ANP \cite{kim2018attentive}, and the ConvCNP \cite{Gordon2020Convolutional} on the in-domain target set of data generated by the EQ and periodic kernels. A noteworthy point is the suboptimal performances of the NP compared to the CNP given that the former learns to estimate the variance in the latent space instead of the output space. Furthermore, inducing shift equivariance bias can be seen to favor ConvCNP even on the in-domain target points. 
\begin{table*}[h!]
    \centering
    \small
     \caption{Mean and standard deviation (over 5 runs) of log likelihood and reconstruction error on the 1-d regression tasks for exponential and periodic function families.}
    \begin{tabular}{c c c c c}
        \toprule
        \multirow{2}{*}{\bfseries Model} & 
        \multicolumn{2}{c}{\bfseries EQ} & 
        \multicolumn{2}{c}{\bfseries Periodic}\\ \cmidrule(lr){2-3} \cmidrule(lr){4-5}
        & NLL & MSE & NLL & MSE \\ \cmidrule(lr){1-5}
        CNP & $1.11 \pm 0.47$ & $0.0204 \pm 0.021$  & $0.199 \pm 0.248$ & $0.052 \pm 0.022$ \\
        NP & $1.087 \pm 0.426$ & $0.02 \pm 0.019$ & $0.172 \pm 0.214$ & $0.052 \pm 0.02$\\
        ANP & $1.108 \pm 0.44$ & $0.0195 \pm 0.021$ & $1.047 \pm 0.585$ & $0.032 \pm 0.023$ \\
        ConvCNP & $2.44 \pm 0.867$ & $0.01 \pm 0.018$ & $1.41 \pm 0.77$ & $0.027 \pm 0.024$ \\
        \bottomrule
    \end{tabular}
    \label{table:1dscores}
\end{table*}

\begin{table}[h!]
\centering
 \caption{Test set log likelihood scores for image completion on MNIST and CelebA-32 datasets.}
\begin{tabular}{@{} l *2c @{}}
\toprule
 \multicolumn{1}{c}{\bfseries Model}    &   \multicolumn{1}{c}{\bfseries MNIST}  &  \multicolumn{1}{c}{\bfseries CelebA-32}    \\ 
\midrule
 CNP & 2062.849 & 2559.732   \\ 
 NP & -2175.749 & 1508.08   \\
 ANP & 2507.989 & 5238.15  \\
GBCoNP & 2743.51 & 5315.59  \\ \bottomrule
 \end{tabular}
 \label{table:testsetscore2d}
\end{table}

\begin{figure}[t!]
    \centering
\begin{minipage}{.8\linewidth}
      \centering
      \includegraphics[width=\textwidth]{figures/experiments/CelebA32.png}\\
\end{minipage}\vfill
\begin{minipage}{.8\linewidth}
        \centering
\includegraphics[width=\textwidth]{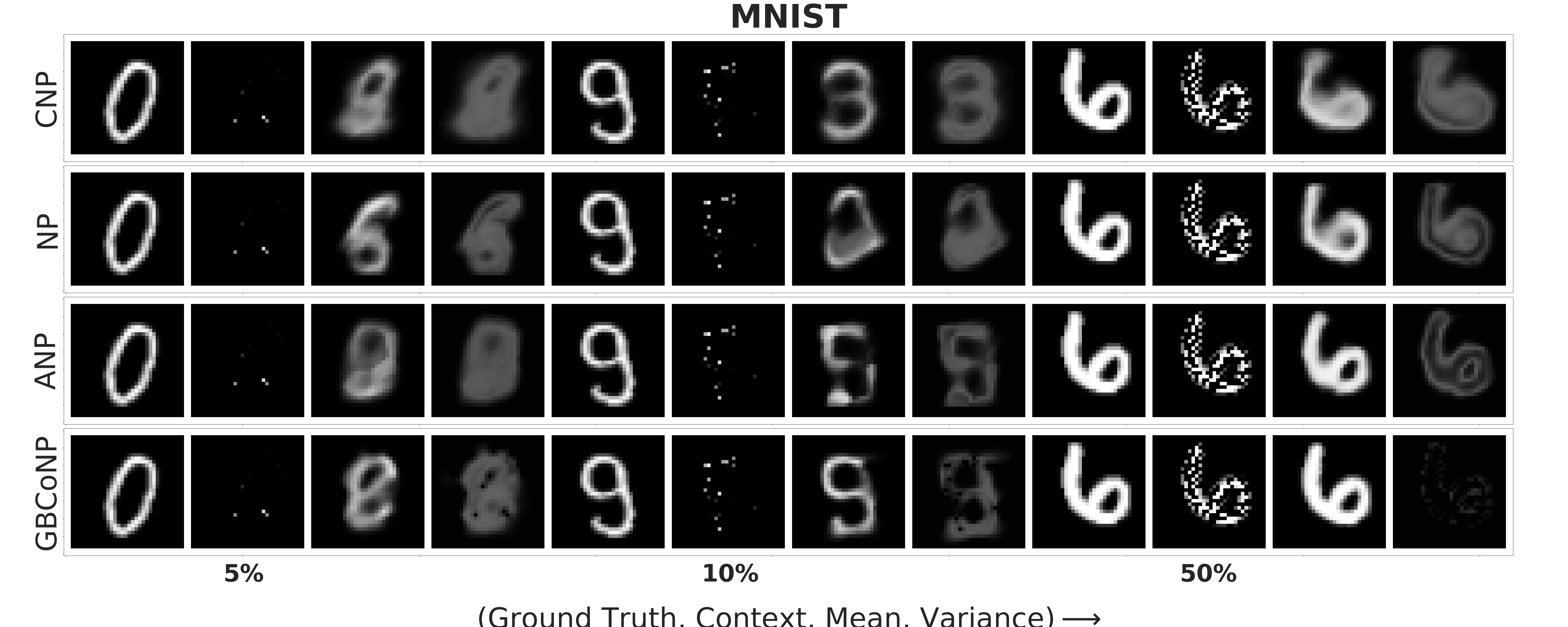}
\end{minipage}
\caption{Predictive mean, and variances of CelebA $32 \times 32$ (top) and MNIST $28 \times 28$ (bottom) images with three different context settings amounting to  5, 10, and 50\% of the total image pixels.}
\label{app:imagecompletion}
\end{figure}
\subsection{2-d function modeling}
\label{sec:image_completion}
For inputs located in 2-d space, we consider modeling the visually intelligible outputs of  $f$ guiding the process of RGB and grayscale image pixel generation.  Modeling such image completion amounts to learning the mapping $f$ between the 2-d input coordinates and their respective output intensities $I$, \textit{i.e.,} $I = [0,1]$ for grayscale and $I = [0,1]^3$ for RGB images \cite{garnelo2018conditional}. We  employ the MNIST handwritten digit \cite{lecun1998gradient} and the CelebA dataset \cite{liu2015faceattributes} to denote grayscale and RGB ranges, respectively. We use the same four NP variants as in 1-d regression task except for ConvCNPs which we replace with its latent variable counterpart GBCoNP \cite{wang2021global} for crisper visualizations. We leverage the pretrained weights of \cite{dubois2020npf} for the CNP, the NP, and the ANP, and that of \cite{wang2021global}  for the GBCoNP.\footnote{\url{https://github.com/xuesongwang/global-convolutional-neural-processes}} Table \ref{table:testsetscore2d} reports the test set log likelihood scores for each model after 50 training epochs.

We visualize the predictive mean and variances of each model by randomly selecting $5\%, 10\%,$ and $50\%$ of the total image pixels as the  context points and providing the locations of the entire image pixels  as targets. As shown in Fig. \ref{app:imagecompletion}, the mean predictions of each model on both the datasets become less blurry with an increase in the context sizes.  In the particular case of Celeb-A dataset with a majority of female faces, the CNP with a lack of latent sampling has its mean predictions resemble closely to the average of all the faces, \textit{i.e.,} more feminine attributes. In terms of variances, an increase in context leads to smoothing of the edges and boundaries that the model is initially more uncertain of. The variances are more pronounced on MNIST where, a context of $5\%$ implies insufficient information and hence, all the models have higher variances throughout the plausible digit surface area.

\begin{figure}[h!]
    \hspace{-4mm}
    \begin{minipage}{0.515\linewidth}
        \centering
        \includegraphics[width=\textwidth]{figures/experiments/CelebA32_img1165zdim23.png}\\
        CelebA-32
    \end{minipage} \vline\hfill
    \begin{minipage}{0.48\linewidth}
        \centering
        \includegraphics[width=\textwidth]{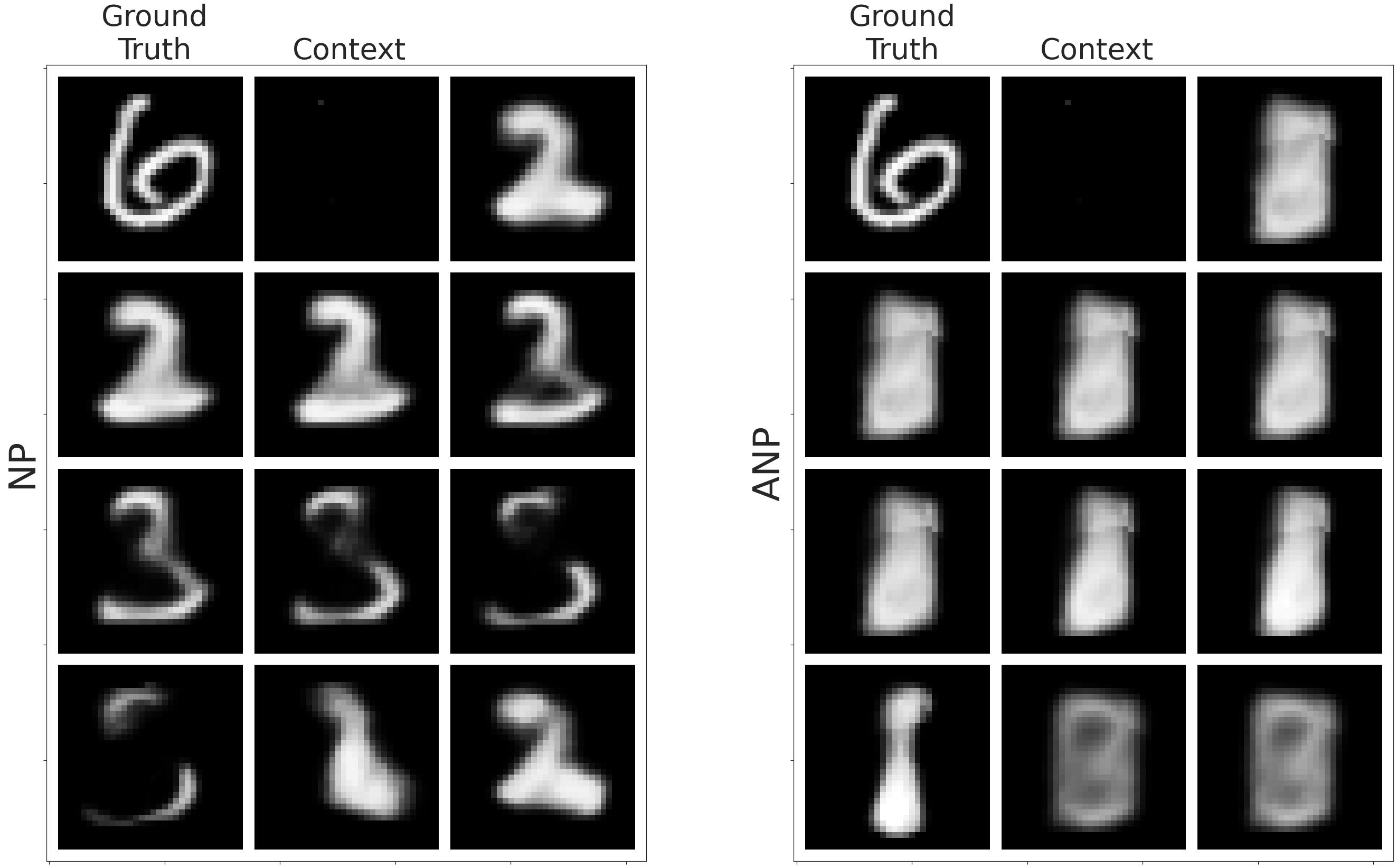}\\
MNIST
    \end{minipage}
    \caption{$4 \times 3$ grids depicting the variation in predictive means of the NP and the ANP over 10 distinct priors on CelebA-32 (left) and MNIST (right) context points, each prior is a sampled latent variable. For reference, the first two  images of each grid represent the ground truth and the context, respectively.}
    \label{app:globaluncertainty}
\end{figure}

To emphasize the role of latent distribution in capturing global uncertainty, we experiment with drawing  samples that are coherent with the observations. To do so, we fix the number of context points and show how latent variable models, namely the NP and the ANP can exploit the covariance between these as well as the target points to generate a range of  predictions that are equally justifiable to the context. Fig. \ref{app:globaluncertainty} shows a few such coherent samples for the models on the CelebA and MNIST datasets. It is worth noting that the deterministic nature of the CNP limits it from generating such coherent possibilities.  In the lack of a latent distribution, sampling from the CNP will amount to mere noises added on top of the model's mean predictions.

\subsection{3-d function modeling}
\label{app:3d_modeling}
To demonstrate an application of NPs for modeling processes involving 3d inputs, we tackle the problem of part labeling on ShapeNet part dataset \cite{yi2016scalable}. Part labeling further shows the application of the NPF for classification tasks.  The ShapeNet part dataset hosts 50 such part labels on 16,881 3d point clouds belonging to 16 categories of objects. Each label is semantically consistent across the shapes of a category.  We adhere to the official train-validation-test split \cite{chang2015shapenet}. Our NP architecture involves incorporating a graph neural network, namely the dynamic graph CNN (DGCNN) \cite{wang2019dynamic}. The input setting follows the segmentation task convention: a list of coordinate locations/features  $(x, y, z)$ capturing the local geometry of an object and an 1d categorical descriptor of the object for the cloud's global representation. Given the part labels of the context set, the goal is to predict the target set labels as well as uncertainties within the same cloud.

 \begin{figure*}[t!]
    \centering
    \includegraphics[width=\textwidth]{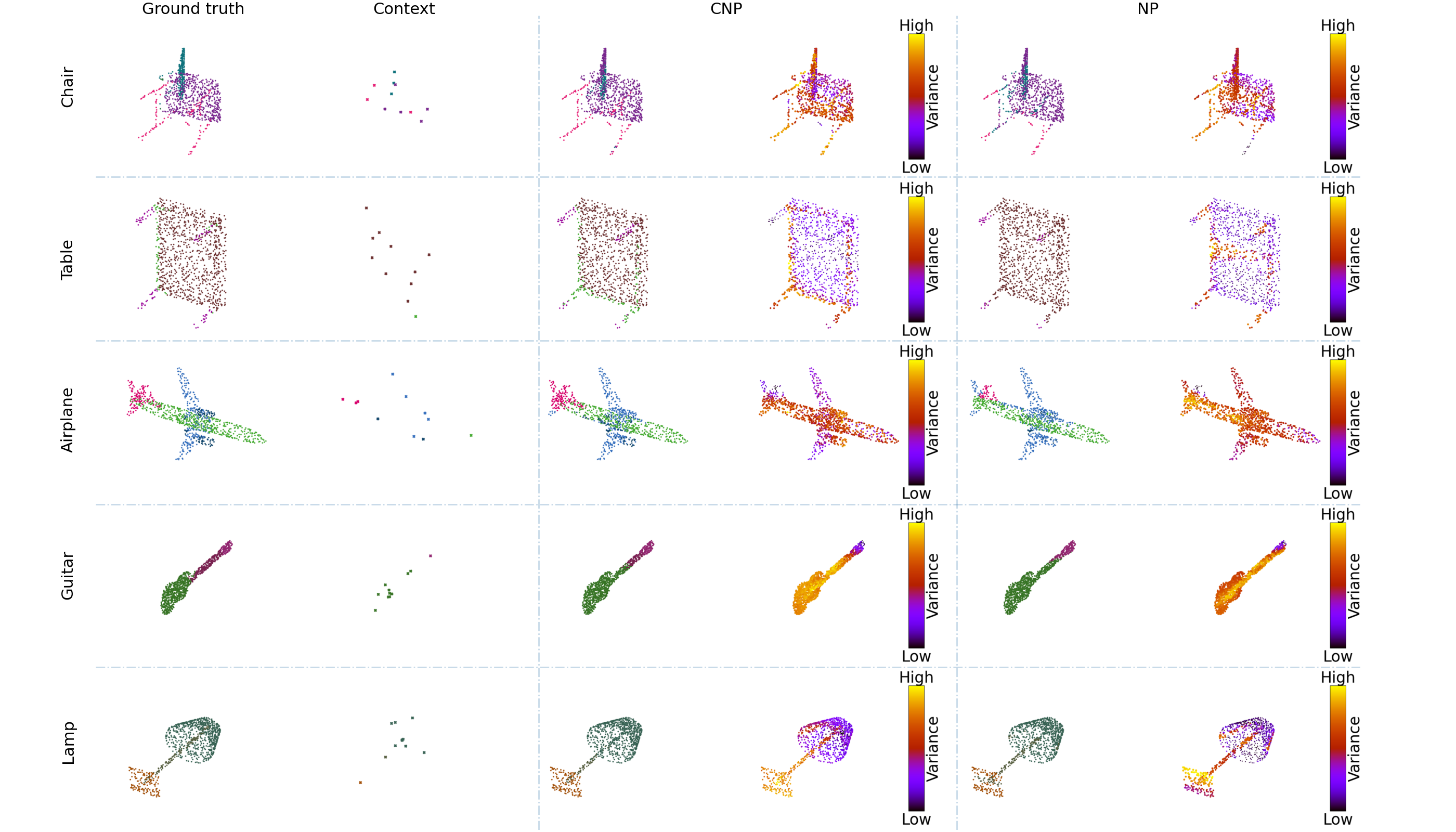}
    \caption{Visualization of part labeling on the ShapeNet part dataset. Each shape depicts 1024 sampled points which form (from left to right): ground truth part labels, 1\% context labels, and mean and variance of the DGCNN-based CNP and NP predictions. For better visibility: (i) the size and color of the points  denote uncertainties, (ii)  the context  points have been magnified  and might differ in spatial alignment with the rest of the shapes.} 
    \label{app:pointcloud}
\end{figure*}
We tailor the original architecture of DGCNN into the encoder and decoder modules of NPs. The encoder extracts the coordinate features of the context set with  edge convolutions. The resulting vector after max-pooling becomes the global representation of the cloud object. For the CNP, we can utilize this representation as the functional prior to decode labels for the target set, whereas for latent NP, such a functional prior parameterizes the latent distribution. 
The inputs to the decoder include the global representation from the encoder, the intermediate edge convolution features extracted from the target set and the categorical descriptor. We modify the final deterministic layer into a Bayesian linear layer (BLL) with the weights and biases having their corresponding means and variances. The prediction is then obtained by sampling the weights and biases from the BLL and calculating the standard cross-entropy loss using these. According to the reparameterization trick \cite{kingma2015variationaltrick}, the uncertainty can be formalized by passing the last hidden layer with the variance of the BLL weights and biases.

Extending NPs to classification setting amounts to eq. \eqref{eq:posterior} defining a categorical distribution  \cite{lukasiewicz2022np}.
Our training objective thus involves minimizing the loss function $\mathcal{L}$ being a linear combination of three components: the categorical cross entropy loss $\mathcal{L}_{CE}$ with label smoothing  \cite{wang2019dynamic}, the normalization of the Bayesian linear layer ($\mathcal{L}_{BKLL}$) and in the case of the latent models, the KL divergence between the prior and posterior distributions ($\mathcal{L}_{KL}$):
\begin{equation}
    \mathcal{L} = \mathcal{L}_{CE}  + \mathcal{L}_{KL} + 0.01* \mathcal{L}_{BKLL}
\end{equation}

\begin{table}[h!]
 \centering
 \footnotesize
 \caption{Mean class accuracy (MCA), overall accuracy (OA), and mean-Intersection-over-Union (mIoU) scores of part labeling  on the ShapeNet test set obtained by fixing the context size to be $1\%$ of the total points}
      \begin{tabular}{@{} l *3c @{}}
\toprule
 \multicolumn{1}{c}{Model}    & MCA & OA & mIoU \\ \midrule
 CNP &  42.103 & 68.705 & 46.388 \\
 NP & 38.823 & 63.247 & 37.645 \\
  \bottomrule
 \end{tabular}
 \label{tab:3d_a}
\end{table}

\begin{table}[h!]
\footnotesize
      \centering
      \caption{Comparison of the performance of the CNP over varying  context sizes.}
\begin{tabular}{@{} l *5c @{}}
\toprule
 \multicolumn{1}{c}{Metric}    & 0.1\%   & 1\% & 5\% & 10\% &  50\% \\ \midrule
 MCA & 42.103  & 47.647 & 61.899 & 63.883 &  64.678 \\ 
 OA & 68.705  & 76.061 & 88.140 & 90.040 &  90.762 \\
 mIoU & 46.388 &  51.589 & 71.894 & 75.907 &  78.348 \\
  \bottomrule
 \end{tabular} 
 \label{tab:3d_b}
\end{table}
We keep our training setup to be the same as that of DGCNN \cite{wang2019dynamic}. Fig. \ref{app:pointcloud} shows
the mean and variance of the DGCNN-based CNP and NP models based on $1\%$ context, \textit{i.e.,} 10 observed points across 5 different categories:  chair, table, airplane, guitar, and lamp. Both the CNP and the NP output more uncertain predictions for points lying at the junctions and boundaries of the shapes. The CNP can be seen to produce more accurate overall predictions than the NP.  The superior prediction results of the CNP is further highlighted in Table \ref{tab:3d_a} where, we compare it against NP in terms of three evaluation metrics: the standard overall accuracy (OA), the mean class accuracy (MCA) computed as the ratio of sum of accuracy for each part predicted to the number of parts, and the mean-Intersection-over-Union (mIoU) score computed as the average of the ratio of overlap, \textit{i.e.}, the true positive score to
the union, \textit{i.e.,} the sum of true positive, false positive and false negative scores for each part. The superior predictive results of the CNP stands in line with the previous works \cite{garnelo2018conditional, garnelo2018neural}.

Given that the DGCNN induces strong graph connectivity bias into the NP, it could be the case that the  performance of NPs have little to do with the observations. To study the importance of the provided context, we ablate the performance of the CNP model by varying the context sizes. In particular, we allow the CNP to take into account a subtotal of $0.1\%,  1\%, 5\%, 10\%,$ and $50\%$ of the sampled points amounting to 1, 10, 51, 102,  and 512 context points, respectively. The number of nearest neighbors $k$ for the former three settings are fixed at 1, 4, and 8, respectively while keeping $k=40$ for the others. The ablation results are shown in  Table \ref{tab:3d_b}. We observe that the amount of observation provided greatly affects the performance of the model across all three metrics. For instance, on the mIoU metric that is commonly used to evaluate segmentation performances, varying context sizes from $0.1\%$ to $50\%$ can help achieve a gain of roughly 32 points.



\ifCLASSOPTIONcaptionsoff
  \newpage
\fi

\end{document}